%% file: main.tex
\documentclass{article} %
\usepackage{iclr2027_conference,times}

\input{misc/math_commands.tex}

\input{misc/commands.tex}
\input{misc/settings}

\newsavebox{\matmulMismatchBox}

\usepackage{hyperref}
\hypersetup{colorlinks=true,linkcolor=blue!50!black,citecolor=blue!50!black,urlcolor=blue!50!black}
\usepackage{url}
\usepackage{graphicx}
\usepackage{float}
\usepackage{xcolor}
\usepackage{amsmath,amsfonts}
\usepackage{soul}
\usepackage{algorithm,algorithmic}
\usepackage{needspace}
\usepackage{wrapfig}
\usepackage{capt-of}

\title{Measuring and Reducing Cross-Vendor Mismatch in Language Models}

\author{Erland Hilman Fuadi\thanks{Corresponding author: \texttt{erland.fuadi@mbzuai.ac.ae}} \quad Chong Tian \quad Xiaosong Ma \quad Qirong Ho \\
Mohamed bin Zayed University of Artificial Intelligence
}

\usepackage{xcolor}

\iclrfinalcopy
\begin{document}

\maketitle
\lhead{Preprint}

\input{sections/abstract}

\input{sections/introduction}

\input{sections/cross_vendor_numerical_compatibility}

\input{sections/method}

\input{sections/experimental_design}

\input{sections/experiment_and_results}

\input{sections/discussion}

\input{sections/related_works}

\input{sections/conclusions}

\newpage
\bibliography{iclr2027_conference}
\bibliographystyle{iclr2027_conference}
\newpage

\appendix

\input{sections/appendix}

\end{document}

%% file: misc/math_commands.tex
\usepackage{amsmath,amsfonts,bm}

\def\eqref#1{equation~\ref{#1}}

\def\1{\bm{1}}

\DeclareMathAlphabet{\mathsfit}{\encodingdefault}{\sfdefault}{m}{sl}
\SetMathAlphabet{\mathsfit}{bold}{\encodingdefault}{\sfdefault}{bx}{n}

%% file: misc/commands.tex
\newcommand{\Method}[1]{ViT Engram}

\newcommand{\profileEqual}{Equal}
\newcommand{\profileMSE}{MSE-heavy}
\newcommand{\profileKL}{KL-heavy}
\newcommand{\profileToken}{Token-heavy}
\newcommand{\profileTopFive}{Top-5-heavy}
\newcommand{\profileNumerical}{Numerical-heavy}
\newcommand{\profileDecision}{Decision-heavy}
\newcommand{\targetbase}{Native AMD (BF16)}

\newcommand{\methodshort}{output-head LoRA}
\newcommand{\resulttodo}[1]{\textcolor{red}{\textbf{[RESULT TODO: #1]}}}

\newcommand{\bitwiseV}{bitwise equality}

\newcommand{\logitsV}{logit differences}

\newcommand{\topkV}{top-$K$ consistency}

\newcommand{\tokenV}{token agreement}

\newcommand{\benchmarkV}{task accuracy}

\newcommand{\reference}{reference}

\newenvironment{draftresulttable}{%
  \small
  \setlength{\tabcolsep}{3pt}%
  \renewcommand{\arraystretch}{1.15}%
  \renewcommand{\resulttodo}[1]{\textcolor{red}{\textbf{TODO}}}%
}{}

\AtBeginDocument{}

%% file: misc/settings.tex
\usepackage{xcolor}
\usepackage{soul}
\usepackage{algorithm}
\usepackage{algorithmic}
\usepackage{capt-of}
\usepackage{wrapfig}
\usepackage{needspace}

\usepackage[textsize=footnotesize]{todonotes}

\definecolor{reviewhighlight}{RGB}{255,242,180}
\sethlcolor{reviewhighlight}

%% file: sections/abstract.tex
\begin{abstract}
Running the same language model on different graphics processing unit (GPU) vendors can produce different logits, even when the model weights and inputs are the same. We analyze cross-vendor mismatch in two dense and two mixture-of-experts (MoE) models with five metric families, namely \bitwiseV, \logitsV, \topkV, \tokenV, and \benchmarkV. We trace one source of the mismatch to accumulation order inside vendors' matrix instructions. Upcasting to FP32 reduces the dense model's logit error by 43\% at three times the runtime, yet keeping only the MLPs in BF16 retains 94\% of this gain at 1.3 times the runtime, so most of the cost of full upcasting buys little. In the MoE models, FP32 and FP16 both lower the probability error but raise the logit error and change expert selection, and FP16 fails in the dense model. An output-head low-rank adapter (LoRA) does not help either, since the final hidden state does not predict the mismatch. The mismatch also carries into training. With every seed fixed, a student distilled from a teacher running on AMD answers 431 MMLU questions differently from one distilled from the same teacher on NVIDIA. Under FP32 upcasting, bitwise equality barely changes while the output distributions move most of the way to the reference, so judging cross-vendor agreement by a single measure misreads both its cost and its gains. Code is available at \url{https://github.com/crova-project/crova}.
\input{figures/fig_hook}
\end{abstract}

%% file: figures/fig_hook.tex
\begin{figure}[H]
    \centering
    \includegraphics[width=\linewidth]{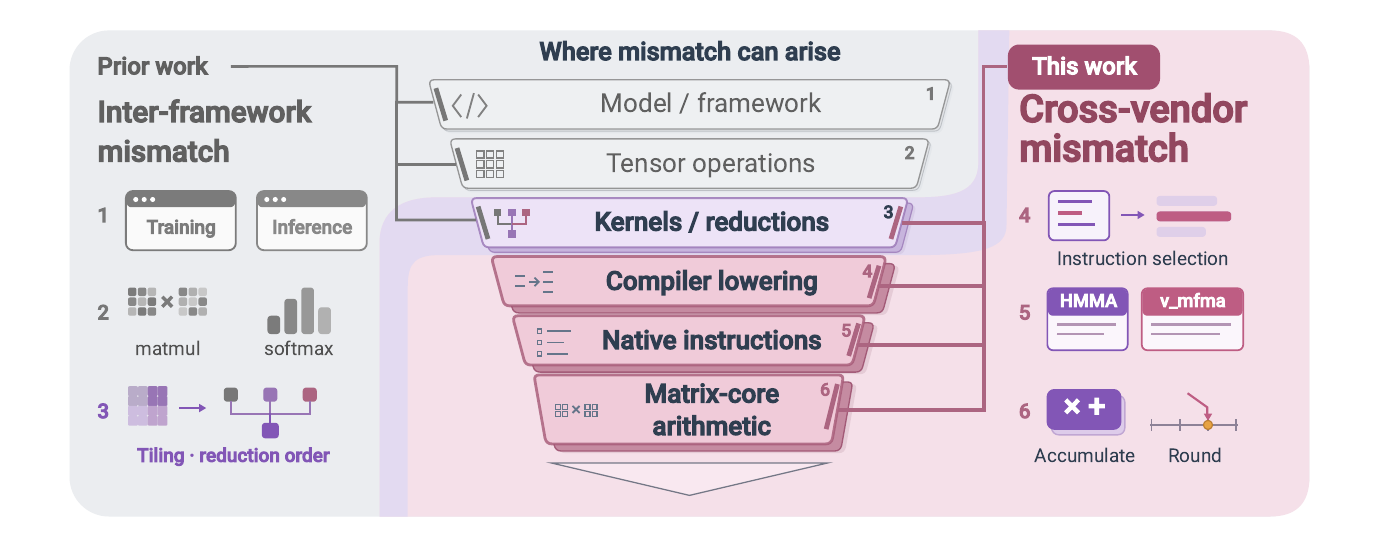}
    \caption{Cross-vendor mismatch spans the execution hierarchy. Inter-framework studies address differences in models, frameworks, and tensor-operation choices. At lower levels, kernel and hardware arithmetic can introduce additional numerical differences. The hierarchy organizes the analysis of how these differences propagate through model outputs and behavior.}
    \label{fig:mismatch-stack}\label{fig:hook}
\end{figure}

%% file: sections/introduction.tex
\section{Introduction}
Large language models (LLMs) are increasingly developed and deployed across heterogeneous hardware and software environments~\citep{chen2025mindvl}. Different execution stacks implement floating-point arithmetic differently, producing different outputs across GPU vendors even when the model weights and inputs are unchanged, which we call cross-vendor mismatch~\citep{li2024fttnfeaturetargetedtestingnumerical,zahid2024testinggpunumericsfinding}. In our measurements, fewer than one in ten output logits of an 8B model are bit-identical between an NVIDIA and an AMD GPU. This matters when a model developed on one vendor is used on another, since its output values determine the probabilities and choices used in later computation. These differences raise questions about the reliability and reproducibility of LLMs across execution environments~\citep{chauvin2026logprobabilitytrackingllm,yuan2025understandingmitigatingnumericalsources}, and assessing them requires examining both the generated answers and the numerical values that produce them.

\input{figures/fig_mismatch}
The consequences of cross-vendor mismatch depend on how these outputs are used. For example, in knowledge distillation (KD), the teacher's probability distribution supplies the student's training targets~\citep{hinton2015distillingknowledgeneuralnetwork}. Differences in these probabilities change the targets even when the teacher selects the same token. Reinforcement learning (RL) also uses output probabilities to compute training objectives, making consistency between generation and training relevant to learning~\citep{qi2025defeatingtraininginferencemismatchfp16, zhong2026diagnosingtraininginferencemismatch}. During generation, a small numerical difference can change the selected token when competing scores are close. In mixture-of-experts (MoE) models, router values determine expert selection and mixture weights, and a changed selection changes which parameters process the token~\citep{ma2025stabilizing}. These choices affect computation and the final answer.

Reducing mismatch requires understanding where the numerical differences arise, from execution frameworks down to hardware arithmetic, as illustrated in \autoref{fig:hook}. Prior work mainly studies inter-framework mismatch in the upper parts of this hierarchy~\citep{ma2025stabilizing,qi2025defeatingtraininginferencemismatchfp16,zhang2026precisiontraininginferencemismatchoptimization,zhong2026diagnosingtraininginferencemismatch}. Other cross-accelerator work validates alternative training stacks against NVIDIA-based frameworks through forward-precision and training-loss comparisons~\citep{chen2025mindvl}. Our work focuses on cross-vendor mismatch at the lower levels of the hierarchy. Here, the underlying matrix arithmetic can differ in its rounding and accumulation behavior across GPU architectures~\citep{li2024fttnfeaturetargetedtestingnumerical}. This is a non-trivial problem that requires deeper investigation at the machine-code level. Matching the model code and framework settings alone therefore does not ensure the same numerical behavior across vendors.

How agreement is measured decides what a study concludes, and prior work measures it incompletely. Deterministic-inference work targets bitwise-identical outputs~\citep{he2025nondeterminism,zhang2025deterministicinferencetensorparallel,yang2026tamingbitwisebehaviorgpu} or identical token sequences~\citep{gond2026llm42enablingdeterminismllm,chu2026margingatesparsemargintriggeredverification}, and LayerCast~\citep{yuan2025understandingmitigatingnumericalsources} judges FP32 upcasting by the spread of accuracy, response length, the first divergent token, and the top-1 probability across NVIDIA configurations only, never across vendors, with its cost reported against full FP32 rather than native BF16. These measures can miss most of what cross-vendor mismatch changes. An intervention can leave bitwise equality, generated tokens, and accuracy almost unchanged while moving the output distribution substantially toward the reference, and an analysis built on those measures alone would conclude that it does nothing. The output distribution is what distillation and reinforcement learning consume, and a cost measured against the wrong baseline hides what an intervention actually costs. Judging cross-vendor agreement therefore requires several metric families at once.

We study cross-vendor mismatch as both an analysis of the disagreement and a comparison of ways to reduce it. We connect internal values, including router values and expert selections in MoE models, to five output-level metric families, and trace the differences arise down to the matrix operations that accumulate them. We then ask where the mismatch can be reduced, by changing the arithmetic format~\citep{qi2025defeatingtraininginferencemismatchfp16,yuan2025understandingmitigatingnumericalsources,cooper2026acceleratingmitigationllminference} or by correcting only the output with a low-rank adapter~\citep{hu2021loralowrankadaptationlarge}, and whether it carries into distillation. We find that the mismatch is widespread and survives the obvious fixes, that most of the gain from higher precision comes without its cost, and that judging agreement by bitwise equality alone misses what matters.

Our contributions are as follows.
\begin{enumerate}
    \item We characterize cross-vendor mismatch in denses and MoE models with five families of agreement measures, from bitwise equality to task accuracy, and trace where it arises by comparing values across layers and isolating accumulation paths in matrix multiplication.
    \item We examine where the mismatch can be reduced, comparing precision changes throughout the forward pass, higher precision in selected modules only, and a low-rank correction at the output head, together with their cost.
    \item We examine how the choice of agreement measure changes the assessment of each intervention, contrasting bitwise equality with the output distributions and choices that downstream uses consume.
    \item We test whether the mismatch carries into downstream training by distilling a teacher run on each vendor and precision into the same student.
\end{enumerate}

%% file: figures/fig_mismatch.tex
\begin{figure}[!t]
    \centering
    \includegraphics[width=0.85\linewidth]{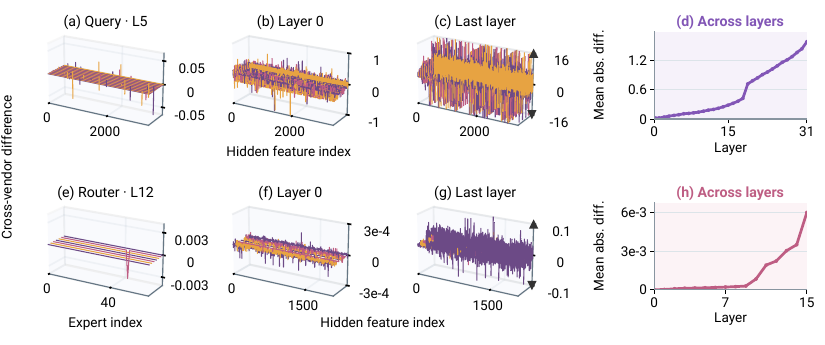}
    \caption{Cross-vendor differences in a Jais dense model \citep[a--d]{anwar2026jais2familyarabiccentric} and  OLMoE MoE model \citep[e--h]{muennighoff2025olmoeopenmixtureofexpertslanguage}. Components (a/e) and first blocks (b/f) receive identical inputs. Last-block outputs (c/g) and layer curves (d/h) use independent full forwards, allowing differences to propagate. The curves show the mean absolute difference across all tokens and hidden features at each block output.}
    \label{fig:mismatch_vendors}
\end{figure}

%% file: sections/cross_vendor_numerical_compatibility.tex
\section{Cross Vendor Numerical Compatibility}
\label{sec:cross-vendor}
\subsection{Mismatch Between Vendors}
Floating-point formats represent values with limited precision. Each arithmetic result is rounded to a representable value, and one unit in the last place (ULP) is the local spacing between them~\citep{goldberg1991floatingpoint}. In 16-bit brain floating point (BF16), one ULP is $0.0078125$ for magnitudes in $[1,2)$ and $0.125$ in $[16,32)$. Execution stacks can change the order of operations and the precision of intermediate results, so a small difference can cross a rounding boundary and then propagate through later layers, even with the same weights and inputs (\autoref{fig:mismatch_vendors}). 

\begin{wraptable}{r}{0.40\linewidth}
\vspace{-\intextsep}
\centering
\caption{Mismatch in the matmul case study.}
\label{tab:matmul-discrepancy}
\input{tables/matmul_discrepancy}
\vspace{-\intextsep}
\end{wraptable}
A controlled matrix-multiplication case study locates one source of these differences (\autoref{sec:kernel-instruction-details}). We compare three accumulation paths for the same projection on both vendors. Vendor general matrix multiplication (GEMM) and a Triton kernel that fixes the order of reduction tiles both differ across vendors at most outputs, and the tiled kernel already differs within its first output tile, where there is no tile order left to control. Sequential scalar accumulation agrees exactly (\autoref{tab:matmul-discrepancy}). The difference therefore lies in how each vendor's matrix instructions accumulate within a tile~\citep{fasi2021numerical}, which standard framework settings do not control. The tiled kernel uses the same matrix instructions as model inference, so any BF16 model on these GPUs inherits this difference.

For an input $x$ and model weights $\theta$, the \reference{} ($R$) and target ($T$) executions produce next-token logits $z_v(x)=f_v(x;\theta)\in\mathbb{R}^{V}$ for $v\in\{R,T\}$, where $V$ is the vocabulary size. Each execution depends on its hardware, software, kernels, precision, and batching, and the reference defines the target outputs. In mixture-of-experts (MoE) models, router values also select each token's experts.

\subsection{Evaluation Metrics}
\label{sec:evaluation_metrics}
\input{figures/fig_evaluation_metrics}
A numerical difference matters through what later computation does with it. A one-ULP change can reorder lower-ranked candidates while the generated answer stays the same, or it can overtake the leading token and change the answer (\autoref{fig:metric-outcomes}). Numerical differences can also change learning targets, such as a teacher's probabilities, when the generated answer agrees.

We therefore define five output-level metric families. \textbf{Bitwise equality} asks whether the executions reproduce exactly the same represented values. \textbf{Logit differences} measure the size of the differences in logits and probabilities. \textbf{Top-$K$ consistency} examines whether the same candidates remain highly ranked and whether their order is preserved. \textbf{Token agreement} compares next-token choices on the same inputs and complete sequences when each execution follows its own continuation. \textbf{Task accuracy} checks whether the resulting answers are correct. For MoE models, we also compare router values and selected experts. \autoref{sec:metric-definitions} gives the complete formulas and evaluation conventions.

%% file: tables/matmul_discrepancy.tex
\begingroup
\footnotesize\centering
\setlength{\tabcolsep}{2pt}
\begin{tabular}{@{}lrr@{}}
\hline
Path & Captured & Synthetic \\
\hline
\multicolumn{3}{@{}l}{\textit{Full projection}} \\
Vendor GEMM & {\scriptsize 14738/15552} & {\scriptsize 14444/15552} \\
Ordered-tile & {\scriptsize 15211/15552} & {\scriptsize 14960/15552} \\
Scalar & {\scriptsize 0/15552} & {\scriptsize 0/15552} \\
\hline
\multicolumn{3}{@{}l}{\textit{First output tile}} \\
Ordered-tile & {\scriptsize 1022/1728} & {\scriptsize 398/1728} \\
Scalar & {\scriptsize 0/1728} & {\scriptsize 0/1728} \\
\hline
\end{tabular}\par\smallskip
\raggedright Unequal / total FP32 elements.\par
\endgroup

%% file: figures/fig_evaluation_metrics.tex
\begin{figure}[!t]
    \centering
    \includegraphics[width=\linewidth]{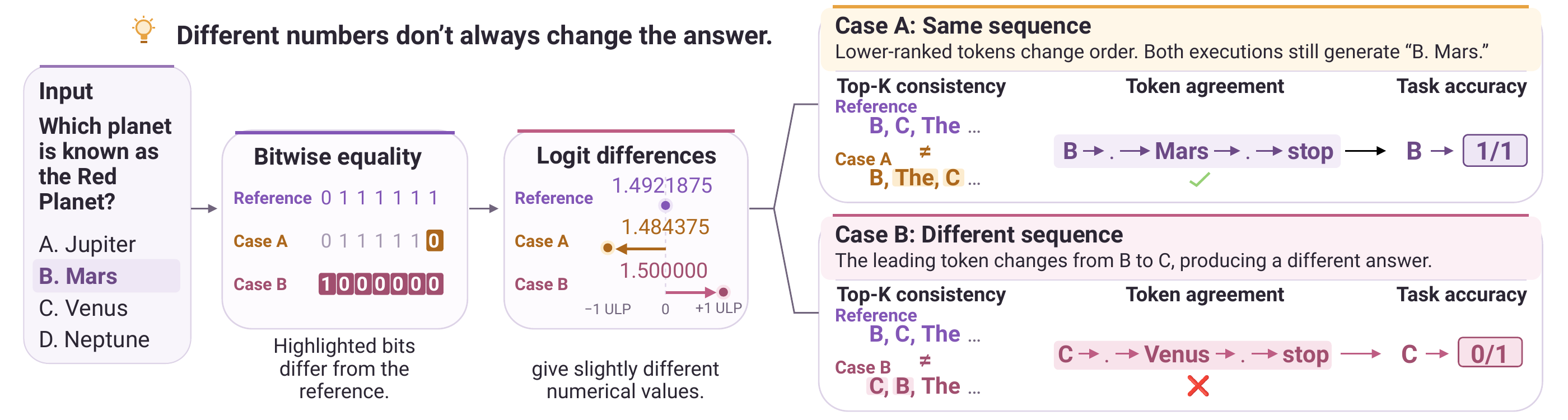}
    \caption{Two constructed examples of numerical mismatch across the five metric families. Case A changes logit values and lower rankings while preserving the generated sequence and correct answer. Case B changes the leading token and leads to an incorrect answer.}
    \label{fig:metric-outcomes}
\end{figure}

%% file: sections/method.tex
\section{Methodology}
\label{sec:method}
We analyze two ways to reduce cross-vendor mismatch against the same fixed NVIDIA reference, each answering one question. First, can the mismatch be mitigated by changing the arithmetic itself? We run the whole forward pass in a different floating-point format, without any training. Second, can the mismatch be represented and corrected at the model's output? We train a low-rank adapter on the output head while the rest of the computation stays unchanged. The two interventions act at opposite ends of the computation, so comparing them shows where the mismatch can be reduced.

\subsection{Precision interventions}
\label{sec:method-precision}
Precision changes act on every operation of the models. The case study in \autoref{sec:cross-vendor} shows one way intermediate values diverge, as the two vendors accumulate the same products in different orders, and \autoref{fig:precision-interventions} shows why the arithmetic format matters for such order differences. Matrix instructions already accumulate BF16 products in FP32, but BF16 execution rounds every intermediate value to BF16 between operations, including the residual stream and normalization outputs. Every partial sum is rounded to the nearest representable value, so a different order changes the result when a partial sum is rounded differently. In the constructed BF16 example, the same three numbers added in two orders end one unit in the last place (ULP) apart, because one order lands exactly halfway between two representable values and the tie is broken in opposite directions. A finer format rounds each partial sum by a smaller amount, so the same order difference produces a smaller discrepancy, and in this example FP16 and FP32 give the same result for both orders. A finer format therefore makes each rounding difference smaller, but it does not remove the different accumulation orders, so it can reduce the mismatch without guaranteeing identical outputs.

\paragraph{FP32 upcasting.} Following LayerCast~\citep{yuan2025understandingmitigatingnumericalsources}, the linear weights stay stored in BF16 while products, activations, partial sums, and logits use 32-bit floating point (FP32). This cannot recover bits absent from the weights, but it keeps more low-order bits in every intermediate value between operations, at the cost of more arithmetic and memory. We test it against the fixed NVIDIA BF16 reference.

\paragraph{FP16.} 16-bit floating point (FP16) has 10 significand and 5 exponent bits, against 7 and 8 for BF16. Its finer grid can absorb small implementation differences at the BF16 memory footprint, as argued for training--inference mismatch in reinforcement learning~\citep{qi2025defeatingtraininginferencemismatchfp16}. Its narrower range risks overflow, so we first check that the outputs are finite.

\input{figures/fig_precision_rationale}

\subsection{Output-head LoRA}
\label{sec:method-lora}
We also test whether a cheap learned correction at the output can absorb the mismatch. The mismatch is small in magnitude, but a small difference is correctable only if it is predictable from what the head receives, so the adapter also serves as a control for whether the mismatch is visible at the output. We add a low-rank adapter $\Delta W = \frac{\alpha}{r}BA$ to the output head $W \in \mathbb{R}^{V \times d}$, with $A \in \mathbb{R}^{r \times d}$, $B \in \mathbb{R}^{V \times r}$, and rank $r \ll \min(V,d)$, and keep every base parameter frozen. For a target hidden value $h_T$, the adapter changes the logits by $\Delta z_T(h_T)=\frac{\alpha}{r}BAh_T$, and losses are evaluated on the corrected native logits. It can change output values, probabilities, and rankings for the same input, but not the hidden states, matrix operations, or expert routing earlier in the forward pass. During generation, a changed token can still change later computation. The adapter therefore tests how much of the mismatch can be corrected at the output alone.

\paragraph{Training objective.}
\label{sec:objective-loss}
We combine four losses that target different aspects of agreement. Raw-logit MSE targets individual values, forward KL divergence the reference distribution, token-gap the leading token's separation from the alternatives, and top-5-gap the gaps that set the order and boundary of the reference top 5 (complete formulas in \autoref{sec:objective-loss-definitions}). Each loss $\ell_j$ is divided by a fixed normalizer $c_j$, its initial mean over training positions, and the objective is $\mathcal{L}=\sum_{j\in\mathcal{J}}w_j\ell_j/c_j$ with $\mathcal{J}=\{\mathrm{MSE},\mathrm{KL},\mathrm{token},\mathrm{top5}\}$. Varying $w_j$ tests which aspects training emphasizes, but a weighting does not guarantee that all agreement measures improve together.

%% file: figures/fig_precision_rationale.tex
\begin{figure}[t]
    \centering
    \includegraphics[width=\linewidth]{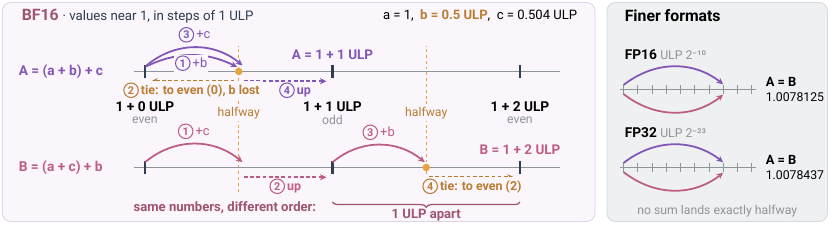}
    \caption{BF16 can gives different value due to less finer on ULP steps. BF16 values near one lie one ULP apart, sums round to the nearest, with ties to the even step. With $b=0.5$~ULP and $c=0.504$~ULP, in $A=(a+b)+c$ \textcircled{\scriptsize 1} $1+b$ lands halfway, \textcircled{\scriptsize 2} rounds down to even so $b$ is lost, and \textcircled{\scriptsize 3}--\textcircled{\scriptsize 4} $c$ rounds it up to $1+1$~ULP. In $B=(a+c)+b$ the sum rounds up to $1+1$~ULP, then $b$ lands halfway and rounds to the even $1+2$~ULP. FP16 and FP32 have finer steps, so both orders agree.}
    \label{fig:precision-interventions}
\end{figure}

%% file: sections/experimental_design.tex
\section{Experimental Design}
\label{sec:experimental-design}
We study Jais 8B~\citep{anwar2026jais2familyarabiccentric} as the dense model and OLMoE~\citep{muennighoff2025olmoeopenmixtureofexpertslanguage} as the MoE model. NVIDIA A100 provides the reference and AMD Instinct MI210 runs the target. NVIDIA A100 uses the Ampere architecture, and AMD Instinct MI210 uses the CDNA2 architecture, which was built as a direct competitor to Ampere. The two therefore belong to a comparable generation. The reference serves as an operational anchor rather than a claim that it is more accurate, so we report reference fidelity and task accuracy seperately.

For Jais, NVIDIA generates greedy reference responses of up to 1,024 tokens, 12.5 tokens long on average, for 1,200 training and 200 development questions, split equally between Massive Multitask Language Understanding (MMLU)~\citep{hendrycks2021measuringmassivemultitasklanguage} and ArabicMMLU~\citep{koto-etal-2024-arabicmmlu}, the natural second benchmark for the Arabic-centric Jais, where the English-only models use ARC-Challenge. AMD hidden values on the training responses train the adapters in an offline fashion. We evaluate on the development responses and on the deduplicated MMLU and ArabicMMLU test sets, which include the adapter's training questions (\autoref{tab:accuracy-no-overlap} excludes them). For OLMoE, NVIDIA generates 1,024-token references for 640 MMLU and ARC-Challenge questions. The base model never emits an end-of-sequence token, so every response reaches this cap, and rollout agreement is therefore not comparable between the two models. Each comparison requires us to move the full-vocabulary logits of every position from one system to the other, about 13\,GB in BF16 for the 128 development responses alone, so a longer cap would make this transfer impractical. The 128 development responses evaluate every intervention, and the other 512 train the adapters, on their first 128 positions, and the distillation students (\autoref{sec:downstream-learning}). Accuracy uses the complete MMLU and ARC-Challenge test sets. Intervals are paired bootstrap intervals over questions, and \autoref{sec:experimental-details} gives the protocol.

%% file: sections/experiment_and_results.tex
\section{Experiments \& Results}
\label{sec:results}
\subsection{Overall comparison}
\label{sec:correction-results}
\autoref{tab:head-results} compares all interventions across the five output families for the dense Jais model and the mixture-of-experts (MoE) OLMoE model. We use \targetbase{} as the baseline. The precision interventions are FP32 upcasting, our adaptation of LayerCast~\citep{yuan2025understandingmitigatingnumericalsources}, and FP16. The output-head interventions are \methodshort{} with \profileEqual{} and a low-rank adaptation (LoRA) trained only with forward Kullback--Leibler (KL) divergence. No intervention improves every family. On Jais, FP32 upcasting reduces the numerical mismatch substantialy, while neither output-head adapter improves on \targetbase{}. On OLMoE, both precision changes lower the probability error but raise the raw-logit error and match fewer top-5 lists. Output-head LoRA fails in both models, so we do not repeat it, and we test the precision interventions on a second dense model, Llama-3.1-8B, and a second MoE model, Qwen1.5-MoE. Over OLMoE's 1,024-token responses, only about a third of rollouts stay identical to the reference even without any intervention.

\input{tables/main_results}

\subsection{Can precision mitigate the mismatch?}
\label{sec:precision-results}
On Jais, FP32 upcasting brings the numerical outputs much closer to the reference. It reduces raw-logit MSE by 42.8\% and forward KL by 37.1\% relative to \targetbase{} (\autoref{fig:discussion-summary}a). It also matches 126 more ordered top-5 lists. The discrete choices change far less, as it matches two fewer leading tokens and one fewer rollout.

\begin{wraptable}{r}{0.43\linewidth}
\vspace{-\intextsep}
\centering
\caption{Cost on one MI210, same workload within each model. $^\ast$Relative to a native run measured alongside (1,054\,s). Native timings vary by about 7\% between runs. $^\dagger$All except MLPs in FP32 (\autoref{sec:selective-main-results})}
\label{tab:precision-cost}
\scriptsize
\setlength{\tabcolsep}{3pt}
\begin{tabular}{@{}lrr@{}}
\hline
& Time (s) & Memory (GB) \\
\hline
\multicolumn{3}{@{}l}{\textit{Jais}} \\
\targetbase{} & 6,111 & 18.7 \\
FP32 upcasting & 19,245 (3.15$\times$) & 47.7 (2.55$\times$) \\
Selective FP32$^\dagger$ & (1.30$\times$) & (2.02$\times$) \\
\multicolumn{3}{@{}l}{\textit{OLMoE}} \\
\targetbase{} & 1,129 & 14.4 \\
FP32 upcasting & 1,405 (1.24$\times$) & 28.7 (1.98$\times$) \\
FP16 & 1,154 (1.02$\times$) & 14.4 (0.99$\times$) \\
Head LoRA (KL)$^\ast$ & 1,066 (1.01$\times$) & 15.1 (1.05$\times$) \\
Head LoRA (Eq.)$^\ast$ & 1,054 (1.00$\times$) & 15.1 (1.05$\times$) \\
\hline
\end{tabular}
\vspace{-\intextsep}
\end{wraptable}
Task accuracy is essentially unchanged. FP32 upcasting answers 19 more MMLU and seven more ArabicMMLU questions correctly than \targetbase{}, and neither change is statistically significant. The numerical gain has a clear cost. It takes 3.15 times as long and uses 2.55 times the peak GPU memory (\autoref{tab:precision-cost}). On Jais, FP16 produced nonfinite outputs, so it has no numerical comparison. FP16 overflows to 65,504, and the Jais MLP activation of block 20 reaches about $1.2\times10^{5}$ in BF16. Whether FP16 is usable therefore depends on the activation of each model.

On OLMoE both formats are finite, and the result is mixed. Relative to \targetbase{}, FP32 upcasting raises raw-logit MSE by 10.7\% but lowers forward KL by 3.8\% and probability MSE by 8.9\%. FP16 raises raw-logit MSE by 16.9\% and forward KL by about 2\%, and lowers probability MSE by 4.4\%. When the outputs rounded to BF16 before the comparison, the KL reductions become increases (\autoref{sec:precision-details}). Both match fewer ordered top-5 lists. First-layer expert agreement falls from 99.5\% to 92.9\% under FP32 upcasting and 92.6\% under FP16 and stays lower on all 16 layers (\autoref{fig:router-layers}).

The two further models confirm this split (\autoref{sec:extension-models}). FP32 upcasting lowers the logit error of the dense Llama model, while both finer formats move Qwen1.5-MoE further from the reference on every logit measure. Higher precision is therefore not a safe default for matching a BF16 reference.

\input{tables/downstream_kd}
\subsection{Selective FP32 closes the gap at a fraction of the cost}
\label{sec:selective-main-results}
\begin{wrapfigure}{r}{0.46\linewidth}
\vspace{-\intextsep}
\centering
\includegraphics[width=\linewidth]{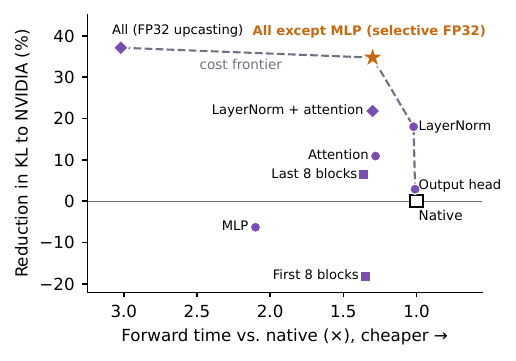}
\caption{Selective FP32 in Jais. The dashed line joins the settings that no cheaper setting beats on KL (\autoref{tab:selective-fp32}).}
\label{fig:selective-frontier}
\vspace{-\intextsep}
\end{wrapfigure}

Full FP32 upcasting, as proposed by LayerCast~\citep{yuan2025understandingmitigatingnumericalsources}, buys its gain with 3.15 times the runtime, so we ask which part of the computation needs the higher precision. \autoref{fig:selective-frontier} upcasts one module type at a time and in combination, with the full results in \autoref{tab:selective-fp32}. Keeping only the MLPs in BF16 retains 94\% of the MSE reduction and 93\% of the KL reduction of full upcasting and matches almost as many top-5 lists, with accuracy unchanged, at 1.30 times the native forward time. It therefore runs 2.3 times faster than full upcasting. The MLPs are the most expensive module to upcast and the least useful, since upcasting them alone raises KL, and Jais is where full upcasting helps most, so even here most of its cost buys little.

The five metric families show why this tradeoff stays invisible to a single measure. Full upcasting raises the share of bit-equal logits only from 9.3\% to 11.2\%, so almost nine in ten values still differ from the reference, yet it lowers KL by 37\% and matches five percentage points more ordered top-5 lists. Closing the gap therefore does not require bitwise equality. Judged by bitwise equality, full upcasting achieves little, and judged by accuracy alone, neither setting changes anything, although the output distributions move most of the way to the reference. These distributions are what distillation and other uses of probabilities consume (\autoref{sec:downstream-learning}), so a system judged on one family can pay for precision it does not need.

\subsection{Can output-head LoRA correct the mismatch?}
\label{sec:lora-results}
Output-head LoRA brings neither model closer to the reference at the common training endpoint. On Jais, \methodshort{} with \profileEqual{} raises raw-logit mean squared error (MSE) and forward KL relative to \targetbase{} and matches fewer bit-equal values and ordered top-5 lists, and the KL-only adapter also raises both and loses rollout agreement. OLMoE gives the same outcome. Both adapters lower their training objective, yet on the development responses they raise forward KL by about 4\% (KL-only $+4.2\%$, \profileEqual{} $+4.1\%$) and raw-logit MSE by 1.7\% and 0.4\%, and they match about 700 fewer ordered top-5 lists out of 131,072. Expert selection is identical to \targetbase{}, because the adapter acts after the last expert layer. The adapters are cheap, costing at most 1\% more time and 5\% more memory (\autoref{tab:precision-cost}), and task accuracy changes by only a few answers. Changing the loss weights does not help either, as all seven weighting profiles end with higher MSE and KL than \targetbase{} in both models (\autoref{sec:detail-loss-balancing}). The adapter fits the training responses but not unseen ones, because the mismatch it must learn behaves like rounding noise rather than a fixed map from the hidden state (\autoref{sec:discussion}).

\subsection{Impact on Downstream Learning}
\label{sec:downstream-learning}
We use knowledge distillation (KD) to test whether cross-vendor mismatch in a teacher changes the student~\citep{hinton2015distillingknowledgeneuralnetwork}. We choose KD because the student learns directly from the teacher's probabilities, so logit differences change its training targets. The teacher is OLMoE and the student is OLMo-1B~\citep{groeneveld2024olmoacceleratingsciencelanguage}, a separate model with the same tokenizer. The student minimizes forward KL to the teacher's 64 most likely tokens, renormalized. Each student learns from one teacher execution, which is NVIDIA BF16, \targetbase{}, AMD FP32 upcasting, or AMD FP16. All students share prompts, reference responses, training setup, starting checkpoint, and AMD hardware, so only the teacher's probabilities differ. A repeated NVIDIA-teacher student shows how much students differ without any teacher mismatch, and a student trained with another data-order seed shows ordinary training variation. \autoref{tab:downstream-kd} compares each student with the NVIDIA-teacher student and reports its accuracy, and \autoref{tab:training-hyperparameters} lists the training settings.

The teacher's mismatch carries into the student. Repeating the NVIDIA-teacher run reproduces the student bit for bit, so any remaining difference comes from the teacher alone. The student distilled from the \targetbase{} teacher differs from the NVIDIA-teacher student by a forward KL of $1.6\times10^{-4}$ and gives a different answer on 431 MMLU questions (3.1\%) and 25 of 1,172 ARC-Challenge questions, while its accuracy changes by only 0.2 points. Upcasting the teacher to FP32 makes this worse, although its own KL to the NVIDIA teacher is the lowest of the three, as its student moves 15 times further in KL and changes 748 MMLU answers (5.3\%). The FP16 teacher behaves like the native one (497 MMLU answers). A different data-order seed changes 2,445 answers of this near-chance student, so the hardware effect is smaller than ordinary training variation. The user chooses a seed, however, while the hardware behind a shared cluster or cloud endpoint is often out of their control, so fixing seeds cannot restore reproducibility.

%% file: tables/main_results.tex
\begin{table}[t]
\providecommand{\secondbest}[1]{\raisebox{0pt}[\height][0pt]{\underline{#1}}}
\centering
\caption{AMD against the NVIDIA reference. Top-5 is ordered top-5 agreement, and rollout the share of responses identical to the reference. Selective FP32 keeps only the MLPs in BF16. Bold is best and underline second best per model. FP16 and FP32 outputs are rounded to BF16 for bit-equal.}
\label{tab:head-results}
\small
\setlength{\tabcolsep}{4pt}
\begin{draftresulttable}
\begin{tabular}{@{}lrrrrrrr@{}}
\hline
Method & \shortstack{Bit-equal\\(\%)$\uparrow$} & \shortstack{Logit\\MSE$\downarrow$} & \shortstack{KL\\($10^{-3}$)$\downarrow$} & \shortstack{Top-5\\(\%)$\uparrow$} & \shortstack{Rollout\\(\%)$\uparrow$} & \shortstack{MMLU\\(\%)$\uparrow$} & \shortstack{AMMLU/\\ARC (\%)$\uparrow$} \\
\hline
\multicolumn{8}{@{}l}{\textit{Jais (dense)}} \\
\targetbase{} & 9.26 & 0.2873 & 7.08 & 66.53 & \secondbest{93.50} & 60.86 & \secondbest{71.36} \\
FP16 & \multicolumn{7}{c}{nonfinite outputs} \\
FP32 upcasting & \textbf{11.24} & \textbf{0.1644} & \textbf{4.45} & \textbf{71.58} & 93.00 & \textbf{61.00} & \textbf{71.40} \\
Selective FP32 & \secondbest{11.07} & \secondbest{0.1717} & \secondbest{4.62} & \secondbest{71.38} & \textbf{94.00} & \secondbest{60.98} & \secondbest{71.36} \\
Head LoRA (KL) & 9.22 & 0.2888 & 7.67 & 65.89 & 92.50 & 60.88 & 71.35 \\
Head LoRA (\profileEqual{}) & 8.65 & 0.2957 & 7.21 & 66.37 & \secondbest{93.50} & 60.87 & 71.34 \\
\hline
\multicolumn{8}{@{}l}{\textit{OLMoE (MoE)}} \\
\targetbase{} & \textbf{14.17} & \textbf{0.0215} & \secondbest{0.194} & \textbf{69.57} & \secondbest{34.38} & 47.37 & \secondbest{54.10} \\
FP16 & 12.85 & 0.0251 & 0.198 & 66.66 & \textbf{35.16} & \textbf{47.46} & \secondbest{54.10} \\
FP32 upcasting & 13.79 & 0.0238 & \textbf{0.187} & 67.75 & 33.59 & \secondbest{47.44} & \textbf{54.69} \\
Head LoRA (KL) & 13.03 & 0.0219 & 0.202 & 69.00 & 33.59 & 47.37 & \secondbest{54.10} \\
Head LoRA (\profileEqual{}) & \secondbest{13.92} & \secondbest{0.0216} & 0.202 & \secondbest{69.04} & 33.59 & 47.37 & \secondbest{54.10} \\
\hline
\multicolumn{8}{@{}l}{\textit{Llama-3.1-8B (dense)}} \\
\targetbase{} & 9.52 & \secondbest{0.0053} & 0.218 & \secondbest{80.27} & \textbf{16.41} & \secondbest{67.04} & \textbf{83.62} \\
FP16 & \secondbest{9.77} & 0.0059 & \secondbest{0.151} & 79.33 & \secondbest{10.94} & \textbf{67.06} & \secondbest{83.53} \\
FP32 upcasting & \textbf{11.50} & \textbf{0.0045} & \textbf{0.130} & \textbf{81.00} & 8.59 & \textbf{67.06} & 83.45 \\
\hline
\multicolumn{8}{@{}l}{\textit{Qwen1.5-MoE (MoE)}} \\
\targetbase{} & \textbf{5.64} & \textbf{0.0152} & \textbf{1.66} & \textbf{66.75} & \secondbest{0.78} & 60.74 & 76.54 \\
FP16 & 4.97 & 0.0199 & 2.30 & 62.22 & \textbf{2.34} & \textbf{60.92} & \textbf{76.79} \\
FP32 upcasting & \secondbest{5.43} & \secondbest{0.0177} & \secondbest{2.01} & \secondbest{63.89} & \textbf{2.34} & \secondbest{60.84} & \secondbest{76.62} \\
\hline
\end{tabular}
\end{draftresulttable}
\end{table}

%% file: tables/downstream_kd.tex
\begin{table}[t]
\centering
\caption{Distilling OLMoE into OLMo-1B with the teacher run on each execution. Same ans.\ is the share of MMLU questions answered as the NVIDIA-teacher student does, and teacher KL is the teacher's own distance from the NVIDIA teacher. Gray numbers give the accuracy change from the NVIDIA-teacher student, and bold marks the best AMD teacher (smallest change for accuracy). The other-seed student changes only the data order.}
\label{tab:downstream-kd}
\small
\setlength{\tabcolsep}{4pt}
\begin{draftresulttable}
\providecommand{\dacc}[1]{\,{\scriptsize\textcolor{gray}{#1}}}
\begin{tabular}{@{}lrrrrrr@{}}
\hline
& \multicolumn{3}{c}{vs.\ NVIDIA-teacher student} & \multicolumn{2}{c}{Accuracy (\%)$\uparrow$} & Teacher \\
\cline{2-4}\cline{5-6}
Teacher & KL ($10^{-4}$)$\downarrow$ & Top-1 (\%)$\uparrow$ & Same ans.\ (\%)$\uparrow$ & MMLU & ARC & KL ($10^{-4}$)$\downarrow$ \\
\hline
No distillation & -- & -- & 20.77 & 24.79\dacc{$-$1.14} & 26.37\dacc{$-$0.51} & -- \\
NVIDIA, repeat & 0 & 100.00 & 100.00 & 25.93 & 26.88 & 0 \\
NVIDIA, other seed & 73.4 & 99.15 & 82.59 & 25.67\dacc{$-$0.26} & 28.50\dacc{$+$1.62} & 0 \\
\targetbase{} & 1.61 & 99.84 & \textbf{96.93} & \textbf{26.12}\dacc{$+$0.19} & 27.22\dacc{$+$0.34} & 1.94 \\
AMD FP32 upcasting & 23.7 & 99.51 & 94.67 & 25.64\dacc{$-$0.29} & 26.79\dacc{$-$0.09} & \textbf{1.87} \\
AMD FP16 & \textbf{1.26} & \textbf{99.87} & 96.46 & 26.21\dacc{$+$0.28} & \textbf{26.88}\dacc{$\pm$0.00} & 1.98 \\
\hline
\end{tabular}
\end{draftresulttable}
\end{table}

%% file: sections/discussion.tex
\section{Discussion}
\label{sec:discussion}

\paragraph{What precision fixes, and what it does not.}
FP32 upcasting acts on the rounding of every intermediate value, but the different accumulation orders remain, so the outputs stay unequal. In OLMoE a finer format removes rounding differences between vendors but also stops rounding the way the BF16 reference does, and a changed expert choice then carries the difference into every later layer. Precision on one vendor alone cannot reach a BF16 reference, yet FP32 on both makes 98.4\% of OLMoE responses identical (\autoref{sec:matched-format}).

\paragraph{Why output-head LoRA is not enough.}
The output head acts after all earlier computation, so it can only map the AMD hidden value that reaches it into a correction. Out of fold, the best linear map from that value increases the error by 0.81\% (\autoref{sec:head-capacity}), and every adapter and weighting ends with higher MSE and KL in both models. Computing the AMD head in FP32 and rounding its output to BF16 reproduces the native logits exactly (\autoref{sec:selective-fp32}), so the difference already sits in the hidden value, in a form a head correction cannot learn.

%% file: sections/related_works.tex
\section{Related Work}
Cross-vendor numerical differences arise at several levels. \citet{li2024fttnfeaturetargetedtestingnumerical} find rounding and accumulation differences across matrix accelerators, and \citet{zahid2024testinggpunumericsfinding} trace program disagreements to mathematical functions and compiler optimizations \citet{yuan2025understandingmitigatingnumericalsources} study variability in LLMs across configurations and propose LayerCast, which keeps BF16 weights and computes in FP32. Another work uses FP16 computation and found that it can reduces training--inference differences in reinforcement learning~\citep{qi2025defeatingtraininginferencemismatchfp16}. Those past work motivate us to do further analysis on the impact of precision towards the LLM outputs for the cross-vendor settings. To lower level, \citet{bhola2024deterministicprobabilisticroundingerror} define that the nondeterminism caused by randomness can be assumed of a probabilistic error model and the way to fix this is to freeze input, execution order, and rounding mode. This is similar with what we found with our experiment on scalar matmul in \autoref{tab:matmul-discrepancy}.

Controlling execution can reduce mismatch. Batch-invariant kernels keep a request's reduction order independent of its batch~\citep{he2025nondeterminism}, fixed reduction trees preserve accumulation order across tensor-parallel sizes~\citep{zhang2025deterministicinferencetensorparallel}, and \citet{yang2026tamingbitwisebehaviorgpu} reconstruct a vendor library's reduction order to reproduce its tensor-core results bit for bit, which makes runs repeatable within one vendor but not identical across vendors. These approaches require control over every kernel on both sides, whereas we instead measure how far unmodified executions diverge and what model-level and precision-level interventions recover. Note that also the past work investigating on the same-vendor, not different vendors like our work. Routing replay stabilizes MoE training~\citep{ma2025stabilizing} which motivates us to investigate the impact on the MoE models. LLM-42 verifies tokens with fixed-size batches~\citep{gond2026llm42enablingdeterminismllm} and MarginGate uses token margins to reduce verification~\citep{chu2026margingatesparsemargintriggeredverification}. Such sequence agreement leaves open how closely the probabilities agree, which we measure alongside generation with our metrics and correct with LoRA~\citep{hu2021loralowrankadaptationlarge}.

%% file: sections/conclusions.tex
\section{Conclusions}
We provide insight on this cross-vendor mismatch across the whole stack, from the accumulation order inside matrix instructions to expert routing, generated answers, and distilled students. The obvious fixes fall short. FP32 upcasting helps the dense model at three times the runtime, and keeping only the MLPs in BF16 retains almost all of this gain at 1.3 times the runtime, but precision gives the MoE models a mixed result, and output-head LoRA cannot reach the computation where the differences arise. A teacher on a different vendor changes 431 of a student's MMLU answers even when training is exactly reproducible, making hardware a hidden variable in distillation and in comparisons across machines. Cross-vendor agreement should therefore be tested like any other model property, and our five families of measures provide the means, since each of these findings appears in some families and not in others. The results also show where the solution lies. Making the accumulation order explicit and shared across vendors in kernels and matrix instructions would remove a major source of the mismatch.

%% file: sections/appendix.tex
\input{sections/metric_definitions}
\input{sections/objective_loss_definitions}
\input{sections/experimental_details}
\input{sections/appendix_sources}

\input{sections/appendix_interventions}

\input{sections/appendix_extensions}
\input{sections/limitations}

%% file: sections/metric_definitions.tex
\section{Metric definitions}
\label{sec:metric-definitions}
\subsection{Output comparisons}
The main text explains the role of each metric family. Here we define how
the measurements are calculated. Let
$\mathcal X=(x_1,\ldots,x_N)$ be the evaluated reference inputs, with each
position counted separately. For reference execution $R$ and target execution $T$, write $z_v(x_i)\in\mathbb{R}^{V}$ for the logits over a vocabulary of size $V$ and $z_{v,ij}=z_{v,j}(x_i)$ for token index $j$, with $v\in\{R,T\}$. Let $\mathbf{1}[\cdot]$ be one when its condition holds
and zero otherwise. All comparisons use the native outputs, and adaptation
replaces $z_T$ with the corrected target output. Agreement counts are divided
by their stated denominators to obtain fractions, or multiplied by 100 after
division to obtain percentages. For mixture-of-experts (MoE) models, we also define the comparison
of router values and selected experts below.
\autoref{fig:metric-outcomes} shows why several families are needed, using two constructed cases for the same question. Candidate C's logit changes by one ULP in each case. In Case A, lower-ranked candidates exchange positions while B remains the leading token, and both executions answer correctly. In Case B, C overtakes B and the answer becomes incorrect. Bitwise equality flags both cases, while the other families separate them.

\paragraph{Bitwise equality.}
Let $b_{v,ij}$ be the bit representation of the native 16-bit brain
floating-point (BF16) value
$z_{v,ij}$. The equal-element count is
\begin{equation}
 C_{\mathrm{bit}}=\sum_{i=1}^{N}\sum_{j=1}^{V}
   \mathbf{1}[b_{T,ij}=b_{R,ij}].
 \label{eq:bitwise-metrics}
\end{equation}
Its fraction is $F_{\mathrm{bit}}=C_{\mathrm{bit}}/(NV)$, and full-tensor equality is $B_{\mathrm{all}}=\mathbf{1}[C_{\mathrm{bit}}=NV]$.
This compares bit patterns, so positive and negative zero are distinct.
Full-tensor equality requires every compared element to match.
This native BF16 measure is not applicable when an execution returns logits
in another format, as in the LayerCast comparison.

\paragraph{Logit differences.}
Raw-logit mean squared error (MSE) measures the average squared difference
between vocabulary values,
\begin{equation}
 \mathrm{MSE}_{z}=\frac{1}{NV}\sum_{i=1}^{N}\sum_{j=1}^{V}
 (z_{T,ij}-z_{R,ij})^2.
 \label{eq:mse}
\end{equation}
At temperature one, the corresponding probabilities are
$p_{v,ij}=\exp(z_{v,ij})/\sum_{a=1}^{V}\exp(z_{v,ia})$.
Probability MSE compares these distributions coordinate by coordinate,
\begin{equation}
 \mathrm{MSE}_{p}=\frac{1}{NV}\sum_{i=1}^{N}\sum_{j=1}^{V}
 (p_{T,ij}-p_{R,ij})^2.
 \label{eq:probability-mse}
\end{equation}
Forward Kullback--Leibler (KL) divergence compares the target distribution
with the reference distribution that supplies the training signal,
\begin{equation}
 \mathrm{KL}_{R\to T}=\frac{1}{N}\sum_{i=1}^{N}\sum_{j=1}^{V}
 p_{R,ij}\bigl(\log p_{R,ij}-\log p_{T,ij}\bigr).
 \label{eq:forward-kl}
\end{equation}
The KL sum over the vocabulary is averaged over inputs. MSE also divides
by the vocabulary size. We use natural logarithms and compute the
probabilities through log-softmax. These quantities respond differently to
the same correction. For example, a shared shift $z_T=z_R+c\mathbf{1}$
gives raw-logit MSE of $c^2$ and leaves both probability measures at zero.

\paragraph{Top-$K$ consistency.}
Let $g_v(x_i)=\operatorname*{arg\,max}_j z_{v,ij}$ be the leading token.
Let $\pi_v^K(x_i)$ be the ordered tuple returned by the execution's top-$K$
selection operation, and let $S_v^K(x_i)$ contain its token indices.
We compare the leading token, the complete ordered list, and its membership as
\begin{equation}
 \begin{aligned}
 C_1&=\sum_{i=1}^{N}\mathbf{1}[g_T(x_i)=g_R(x_i)],\\
 C_{\mathrm{ord},K}&=\sum_{i=1}^{N}
   \mathbf{1}[\pi_T^K(x_i)=\pi_R^K(x_i)],\\
 C_{\mathrm{set},K}&=\sum_{i=1}^{N}
   \mathbf{1}[S_T^K(x_i)=S_R^K(x_i)].
 \end{aligned}
 \label{eq:topk-agreement}
\end{equation}
All three counts have denominator $N$, and the ordered and set comparisons use
$K=5$. Set agreement requires all five members to match. Leading-token
selection uses argmax, while the two top-5 comparisons use the native
top-$K$ operation. These operations retain their recorded tie handling,
as specified in \autoref{sec:implementation-details}.

\paragraph{Token agreement.}
Same Greedy Token compares the next token at each reference input, giving $C_{\mathrm{greedy,fixed}}=C_1$, the same count as Top-1. To compare generated sequences, let $q_m$
be initial prompt $m$ among $M$ prompts, and let $y_{v,m,t}$ be execution
$v$'s token at step $t$. Before stopping, each execution uses its own inputs,
$y_{v,m,t}=g_v(q_m\Vert y_{v,m,<t})$, where $\Vert$ denotes concatenation.
For rollout agreement in the main table, generation stops at the first
end-of-sequence (EOS) token or the response cap $H=1{,}024$. Let
$s_{v,m}=(y_{v,m,1},\ldots,y_{v,m,L_{v,m}})$ be the resulting token sequence,
where $L_{v,m}\leq H$ and an emitted EOS token is retained. We add no
padding or synthetic EOS token. The matching-sequence count is
\begin{equation}
 C_{\mathrm{greedy},H}=\sum_{m=1}^{M}
 \mathbf{1}[s_{T,m}=s_{R,m}].
 \label{eq:rollout-agreement}
\end{equation}
Its denominator is $M$. Equality requires both sequence lengths and all
generated tokens to match. Capped responses are compared over the observed
horizon, so this measure does not require both sequences to end in EOS.
For comparison, agreement at every valid reference input is
$\sum_{m=1}^{M}\prod_{t=1}^{L_{R,m}}
\mathbf{1}[g_T(x^R_{m,t})=g_R(x^R_{m,t})]$,
where $x^R_{m,t}$ follows the saved reference continuation. This comparison
uses the reference inputs, whereas \eqref{eq:rollout-agreement} follows
each execution's generated continuation. The reference-input comparison ends
at the reference's first EOS token or cap. Each rollout ends at its own
first EOS token or cap.

\paragraph{Task accuracy.}
For a benchmark $\mathcal B=\{(q_m,a_m)\}_{m=1}^{M_{\mathcal B}}$, let $a_m$
be the correct answer and $\widehat a_v(q_m)$ the answer parsed from execution
$v$'s generated response. The correct-answer count is
\begin{equation}
 C_{\mathrm{correct},v}^{\mathcal B}=
 \sum_{m=1}^{M_{\mathcal B}}
 \mathbf{1}[\widehat a_v(q_m)=a_m].
 \label{eq:task-accuracy}
\end{equation}
The accuracy is $\mathrm{Acc}_v^{\mathcal B}=C_{\mathrm{correct},v}^{\mathcal B}/M_{\mathcal B}$. A failed parse counts as incorrect. We report each benchmark separately for
the unadapted and adapted target, counting questions equally. This measures
whether the resulting answers remain correct after the numerical correction.

\subsection{Expert routing}
For MoE models, let $r_v^{i\ell}\in\mathbb R^E$ be the router logits at
evaluated position $i$ and layer $\ell$, with $N_{\mathrm r}$ positions,
$L$ layers, and $E$ experts. Write
$s_v^{i\ell}=\operatorname{softmax}(r_v^{i\ell})$ at temperature one.
Router-logit MSE and router forward KL are
\begin{align}
 \mathrm{MSE}_{\mathrm{router}}
 &=\frac{1}{N_{\mathrm r}LE}
 \sum_{i=1}^{N_{\mathrm r}}\sum_{\ell=0}^{L-1}\sum_{e=1}^{E}
 (r_{T,e}^{i\ell}-r_{R,e}^{i\ell})^2,
 \label{eq:router-mse}\\
 \mathrm{KL}_{\mathrm{router}}
 &=\frac{1}{N_{\mathrm r}L}
 \sum_{i=1}^{N_{\mathrm r}}\sum_{\ell=0}^{L-1}\sum_{e=1}^{E}
 s_{R,e}^{i\ell}(\log s_{R,e}^{i\ell}-\log s_{T,e}^{i\ell}).
 \label{eq:router-kl}
\end{align}

Let $a_v^{i\ell}$ be the ordered tuple of $K_{\mathrm r}$ expert indices
actually selected by the router, and let $S_v^{i\ell}$ be its set of members.
The selection counts are
\begin{equation}
 \begin{aligned}
 C_{\mathrm{route,ord}}&=
 \sum_{i=1}^{N_{\mathrm r}}\sum_{\ell=0}^{L-1}
 \mathbf{1}[a_T^{i\ell}=a_R^{i\ell}],\\
 C_{\mathrm{route,set}}&=
 \sum_{i=1}^{N_{\mathrm r}}\sum_{\ell=0}^{L-1}
 \mathbf{1}[S_T^{i\ell}=S_R^{i\ell}],\\
 C_{\mathrm{route,all}}&=
 \sum_{i=1}^{N_{\mathrm r}}\prod_{\ell=0}^{L-1}
 \mathbf{1}[a_T^{i\ell}=a_R^{i\ell}].
 \end{aligned}
 \label{eq:routing-agreement}
\end{equation}
The first two counts use denominator $N_{\mathrm r}L$, while agreement across
all layers uses $N_{\mathrm r}$ and requires the ordered selections to match
at every layer for one position. We use the router's actual selections,
retaining its native tie behavior. For the reference, the probability gap
$s_{R,(K_{\mathrm r})}^{i\ell}-s_{R,(K_{\mathrm r}+1)}^{i\ell}$ between
the last selected rank and the next rank describes how close those choices
are, where parentheses denote descending probability rank. We report its mean
and minimum across position-layer pairs. Router values and expert choices
describe the mismatch inside the model, alongside the output-level measures.

%% file: sections/objective_loss_definitions.tex
\subsection{Objective-loss definitions}
\label{sec:objective-loss-definitions}
\autoref{tab:objective-losses} gives the four loss formulas. For reference, let
$\delta=\hat z_T-z_R$ be the remaining logit error,
$p_R=\operatorname{softmax}(z_R)$, and
$\hat p_T=\operatorname{softmax}(\hat z_T)$ at temperature one. Let
$q_1,\ldots,q_5$ be the five highest-scoring reference tokens, ordered by
decreasing score with ties broken by ascending token index, and let
$Q_5=\{q_1,\ldots,q_5\}$. This appendix records the normalization used for
the weighting experiments.

For $N$ questions with $m_i$ valid positions for question $i$, the fixed
normalizer for component $j$ is its initial mean over valid training positions,
computed from the unadapted target and reference outputs. In full notation,
$c_j=(\sum_i m_i)^{-1}\sum_i\sum_{t=1}^{m_i}
\ell_j(z^{(0)}_{T,it},z_{R,it})$. The same $c_j$ values are used for every
weighting profile.

\begin{table}[ht]
\centering
\caption{Individual loss formulas for one input, before normalization and weighting.}
\label{tab:objective-losses}
\small
\renewcommand{\arraystretch}{2.2}
\begin{tabular}{@{}ll@{}}
\hline
Loss & Formula \\
\hline
Raw-logit MSE & $\displaystyle \ell_{\mathrm{MSE}}=\frac{1}{V}\sum_{j=1}^{V}\delta_j^2$ \\
Forward KL & $\displaystyle \ell_{\mathrm{KL}}=\sum_{j=1}^{V}p_{R,j}\log\frac{p_{R,j}}{\hat p_{T,j}}$ \\
Token-gap & $\displaystyle \ell_{\mathrm{token}}=\frac{1}{V-1}\sum_{j\ne q_1}(\delta_{q_1}-\delta_j)^2$ \\
Top-5-gap & $\displaystyle \ell_{\mathrm{top5}}=\frac{1}{5}\biggl[\sum_{i=1}^{4}(\delta_{q_i}-\delta_{q_{i+1}})^2+\frac{1}{V-5}\sum_{j\notin Q_5}(\delta_{q_5}-\delta_j)^2\biggr]$ \\[1.2ex]
\hline
\end{tabular}
\end{table}

%% file: sections/experimental_details.tex
\section{Experimental details}
\label{sec:experimental-details}

\input{tables/training_hyperparameters}

\subsection{Setup}
\label{sec:implementation-details}
\label{sec:olmoe-protocol}\label{sec:kd-protocol}
Both models run with identical weights, tokenizer, and input tokens on the NVIDIA A100 and the AMD MI210, in BF16 with eager attention on one device, batch size one, no key-value cache during evaluation, and with mixed precision, tensor-float-32 arithmetic, and deterministic modes disabled. Both systems use PyTorch 2.7.1 and Transformers 5.13.0, with CUDA 12.8 on the NVIDIA A100 and ROCm 6.3.3 on the AMD MI210. Jais is a dense model with 32 blocks and a vocabulary of 150,272 tokens, and OLMoE has 16 blocks that each select 8 of 64 experts, with a vocabulary of 50,304 tokens. Questions come from the public MMLU, ArabicMMLU, and ARC-Challenge~\citep{clark2018thinksolvedquestionanswering} partitions, with duplicates removed and development questions chosen before a disjoint training set. Jais is a chat model, so each prompt asks for one answer letter and each prefix is evaluated by its own forward pass. OLMoE is a base model whose responses all reach the 1,024-token cap, so each response is evaluated in one teacher-forced forward that records the logits and router outputs at every position. Repeating a run on the same vendor reproduces every logit bit for bit, as the audited repeat of all 200 Jais development responses on AMD and the repeated NVIDIA-teacher student show, so the cross-vendor differences do not come from run-to-run variation.

\input{tables/accuracy_no_overlap}

\subsection{Output-head LoRA training}
We place the adapter at the output head because it performed best among the placements we tried, which also included full-model LoRA and adapters on the final down-projection and the final MLP. \autoref{tab:training-hyperparameters} lists the hyperparameters. Each proposed update is tried at step fractions from 1 down to 1/64 and kept only if it lowers the objective, and each response is weighted by its number of valid positions so that every token counts equally. The normalizers $c_j$ of \autoref{sec:objective-loss} are computed once from the unadapted target and shared by every weighting profile. OLMoE adapters use the same recipe on the first 128 positions of each training response.

\subsection{Distillation targets}
The student learns from each teacher's 64 most likely tokens, renormalized. Against the NVIDIA teacher, these targets share on average 62.1 of 64 tokens for the \targetbase{} teacher, 62.0 for the FP32 teacher, and 61.9 for the FP16 teacher, with a total-variation distance of about 0.002 in each case. The FP32 teacher's training targets are therefore as close to the NVIDIA targets as the native ones, so the selection of the top 64 tokens does not explain why its student moves further.

\subsection{Precision and cost}
\label{sec:precision-details}
FP32 upcasting follows LayerCast~\citep{yuan2025understandingmitigatingnumericalsources} and keeps the linear weights in BF16 while computing in FP32, and FP16 casts the checkpoint to FP16 for all computation. Runtime is wall-clock time including model loading, and memory is peak allocated GPU memory, both measured over the complete benchmark and development workload on one MI210. Jais FP16 produced nonfinite outputs, so we report FP16 only for OLMoE. FP32 upcasting produces FP32 logits while the reference logits are BF16, so part of any measured difference could come from the reference's own rounding. Rounding the Jais FP32 logits to BF16 before the comparison leaves the gain almost unchanged, with raw-logit MSE $-42.5\%$ and forward KL $-33.7\%$ relative to \targetbase{}. For OLMoE on 32 development responses, rounding leaves the rise in raw-logit MSE unchanged ($+13.5\%$ for FP32 upcasting and $+20.7\%$ for FP16) and turns the small KL reductions into increases ($+5.3\%$ and $+11.5\%$). Once both sides are compared in BF16, the finer formats therefore move OLMoE further from the reference on both measures.

%% file: tables/training_hyperparameters.tex
\begin{table}[ht]
\centering
\caption{Training hyperparameters. The normalizers $c_j$ are the mean unadapted losses over each model's training positions. $^{*}$First 128 tokens of each response.}
\label{tab:training-hyperparameters}
\small
\setlength{\tabcolsep}{3pt}
\begin{tabular}{@{}llll@{}}
\hline
& Jais head LoRA & OLMoE head LoRA & Distillation \\
\hline
Trained weights & head, $r=32$, $\alpha=64$ & head, $r=32$, $\alpha=64$ & all student weights \\
Optimizer & AdamW, $10^{-4}$, $(0.9,0.999)$ & AdamW, $10^{-4}$, $(0.9,0.999)$ & AdamW, $10^{-5}$, $(0.9,0.95)$ \\
Weight decay & 0 & 0 & 0 \\
Batch size & 1 & 1 & 8 \\
Training responses & 1,200 & 512$^{*}$ & 512 \\
Epochs & 2 & 2 & 2 \\
Schedule & constant & constant & cosine, 8 warmup \\
Gradient clipping & none & none & 1.0 \\
\hline
$c_{\mathrm{MSE}}$ / $c_{\mathrm{KL}}$ & 0.266 / 0.00814 & 0.00763 / 0.000638 & -- \\
$c_{\mathrm{token}}$ / $c_{\mathrm{top5}}$ & 2.83 / 0.846 & 0.0181 / 0.0157 & -- \\
\hline
\end{tabular}
\end{table}

%% file: tables/accuracy_no_overlap.tex
\begin{table}[t]
\centering
\caption{Jais task accuracy without the questions used for adapter training or development (13,182 MMLU and 13,602 ArabicMMLU questions). Parser failures count as incorrect.}
\label{tab:accuracy-no-overlap}
\begin{draftresulttable}
\setlength{\tabcolsep}{6pt}
\begin{tabular}{@{}lrr@{}}
\hline
Method & MMLU (\%) $\uparrow$ & ArabicMMLU (\%) $\uparrow$ \\
\hline
\targetbase{} & 61.05 & 71.56 \\
Head LoRA (KL) & 61.06 & 71.56 \\
\multicolumn{3}{@{}l}{\textit{Head LoRA weighting profiles}} \\
\profileEqual{} & 61.05 & 71.55 \\
\profileMSE{} & 61.07 & 71.56 \\
\profileKL{} & 61.06 & 71.55 \\
\profileToken{} & 61.08 & 71.56 \\
\profileTopFive{} & 61.12 & 71.53 \\
\profileNumerical{} & 61.04 & 71.55 \\
\profileDecision{} & 61.09 & 71.56 \\
\hline
\end{tabular}
\end{draftresulttable}
\end{table}

%% file: sections/appendix_sources.tex
\section{Extra Details of Where the mismatch arises}
\label{sec:appendix-sources}

\subsection{Expert routing by layer in OLMoE}
\label{sec:detail-routing}
\autoref{fig:router-layers} gives the router agreement of \autoref{sec:precision-results} layer by layer, comparing the eight experts that AMD and the reference select at each layer and position under the same input. With \targetbase{}, agreement starts at 99.5\% in the first layer and falls with depth to 82--85\% in the last three layers. Both finer formats start lower, at 92.9\% (FP32 upcasting) and 92.6\% (FP16), and stay below \targetbase{} at every layer, so the loss begins at the first router and persists through the model.

We also ask whether these disagreements are tie-breaking artifacts. On the first 32 development responses, we recompute the AMD router values and measure the reference gap between its 8th and 9th experts. Only 2.8\% of first-layer positions have a gap of at most one BF16 step, yet 65\% of the \targetbase{} disagreements fall there, so native disagreements are mostly near-ties broken differently. Under FP32 upcasting and FP16, only 20\% of the much more frequent first-layer disagreements fall at such near-ties. The finer formats therefore change the router inputs by more than tie-breaking, which matches their larger logit error.
\input{figures/fig_router_layers}

\input{sections/kernel_instruction_details}

%% file: figures/fig_router_layers.tex
\begin{figure}[H]
    \centering
    \includegraphics[width=\linewidth]{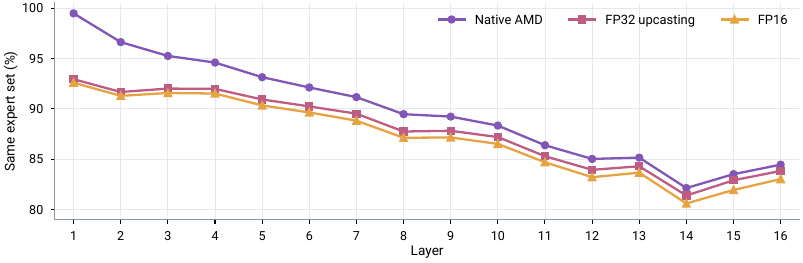}
    \caption{Share of the 131,072 OLMoE development positions at which AMD selects the same eight experts as the NVIDIA reference, by layer. \targetbase{} matches the reference almost exactly at the first layer, while FP32 upcasting and FP16 lose 6.5 and 6.9 percentage points there, and remain below \targetbase{} at every layer.}
    \label{fig:router-layers}
\end{figure}

%% file: sections/kernel_instruction_details.tex
\subsection{Kernel and matrix-instruction case study}
\label{sec:kernel-instruction-details}
This case study supports the accumulation-order account in \autoref{sec:method-precision}. It uses a separate, smaller workload, so it illustrates the mechanism rather than establishing the cause of the model differences in \autoref{sec:results}.

\paragraph{Workload.}
We take the layer-0 query projection of SmolLM2-135M-Instruct ($M=27$, $N=576$, $K=576$) with BF16 inputs and weights and FP32 accumulation and output, on the A100 and the MI210. Each comparison uses the same captured inputs and weights, and is repeated with synthetic inputs. \autoref{tab:matmul-discrepancy} in \autoref{sec:cross-vendor} counts unequal FP32 outputs among 15,552 for three implementations, namely vendor general matrix multiplication (GEMM), a shared Triton kernel with a fixed tile order ($32\times64\times32$ tiles, four warps, two stages, no autotuning, no split reduction), and a scalar Triton kernel that accumulates sequentially over $K$. On AMD, the FP32-output GEMM request falls back from hipBLASLt to another library path.

\paragraph{Results.}
All three implementations are repeatable within each vendor. Vendor GEMM and the ordered-tile kernel differ across vendors, by at most $3.43\times10^{-5}$ and $4.58\times10^{-5}$. The scalar kernel agrees exactly on both inputs. Fixing the outer tile order is therefore not enough, and the remaining difference lies inside the matrix instructions that compute each tile. The compiled MI210 ordered-tile kernel contains the matrix fused multiply-add (MFMA) instruction \texttt{v\_mfma\_f32\_32x32x8bf16\_1k} (32 static occurrences) and reproduces all 15,552 saved outputs bit for bit. The first-tile rows of \autoref{tab:matmul-discrepancy} evaluate only the first $27\times64$ output tile over the reduction range $[0,32)$, as separate one-stage kernels, and show that differences already appear there. Algorithm~\ref{alg:matmul-full} gives the complete procedures.

\input{figures/matmul_full_algorithms}

\paragraph{Instruction probes.}
\autoref{tab:matrix-instructions} lists the instructions in two controlled probes of a logical $32\times32\times16$ operation. On the MI210, two nonzero products of $2^{-24}$ added to an FP32 accumulator of $1$ give $1+2^{-23}$ when placed at reduction positions 0 and 1, but $1$ when placed at positions 0 and 8, one ULP apart. Where a contribution sits within the instruction sequence thus changes the result. On the A100, computing the same operation with width-eight or width-sixteen instructions gives identical outputs on all 28 registered examples and across 100 repetitions, so reduction width alone does not change the A100 result here.

\begin{table}[ht]
\centering
\caption{Matrix instructions in the controlled probes, for the complete logical $32\times32\times16$ operation. Both PTX source names end in \texttt{.row.col.f32.bf16.bf16.f32}.}
\label{tab:matrix-instructions}
\footnotesize
\setlength{\tabcolsep}{3pt}
\begin{tabular}{@{}llll@{}}
\hline
Path & Source instruction & Machine instruction & Count \\
\hline
A100, $K=16$ & \texttt{mma.sync.aligned.m16n8k16}
 & \texttt{HMMA.16816.F32.BF16} & 8 \\
A100, $K=8+8$ & \texttt{mma.sync.aligned.m16n8k8}
 & \texttt{HMMA.1688.F32.BF16} & 16 \\
MI210, $K=8+8$ & \texttt{v\_mfma\_f32\_32x32x8bf16\_1k}
 & \texttt{v\_mfma\_f32\_32x32x8bf16\_1k} & 2 \\
\hline
\end{tabular}
\end{table}

%% file: figures/matmul_full_algorithms.tex
\begin{algorithm}[H]
\caption{Complete matrix-multiplication procedures}
\label{alg:matmul-full}
\small
\noindent Given BF16 matrices $X\in\mathbb{R}^{M\times K}$ and
$W\in\mathbb{R}^{N\times K}$, each path returns an FP32 output
$Y\in\mathbb{R}^{M\times N}$. The explicit kernels use FP32 accumulators.
For ordered-tile matmul, $B_M=32$, $B_N=64$, $B_K=32$, and
$T=\lceil K/B_K\rceil$. Partition $X,W$ into zero-padded tiles
$X_{it}\in\mathbb{R}^{B_M\times B_K}$ and
$W_{jt}\in\mathbb{R}^{B_N\times B_K}$. Output writes retain only indices
inside $Y$.\par\smallskip
\noindent
\begin{minipage}[t]{0.30\linewidth}
\vspace{0pt}
\textbf{(a) Vendor GEMM}\par
\footnotesize
\begin{algorithmic}[1]
\STATE $Y\leftarrow\operatorname{VendorGEMM}(X,W^\top)$
\RETURN $Y$
\end{algorithmic}
The library selects its own implementation and reduction order.
\end{minipage}\hfill
\begin{minipage}[t]{0.34\linewidth}
\vspace{0pt}
\textbf{(b) Ordered-tile matmul}\par
\footnotesize
\begin{algorithmic}[1]
\FORALL{output tiles $(i,j)$}
\STATE $C\leftarrow 0_{B_M\times B_N}$ in FP32.
\FOR{\textcolor{red!70!black}{$t=1,\ldots,T$}}
\STATE Load $A\leftarrow X_{it}$, $B\leftarrow W_{jt}$.
\STATE \textcolor{red!70!black}{$C\leftarrow\operatorname{dot}(A,B^\top,C)$}
\ENDFOR
\STATE Write $Y_{ij}\leftarrow C$.
\ENDFOR
\RETURN $Y$
\end{algorithmic}
\end{minipage}\hfill
\begin{minipage}[t]{0.32\linewidth}
\vspace{0pt}
\textbf{(c) Scalar matmul}\par
\footnotesize
\begin{algorithmic}[1]
\FORALL{output elements $(i,j)$}
\STATE Initialize $c=0$ in FP32.
\FOR{$k=1,\ldots,K$}
\STATE $x\leftarrow\operatorname{FP32}(X_{ik})$
\STATE $w\leftarrow\operatorname{FP32}(W_{jk})$
\STATE \textcolor{red!70!black}{$c\leftarrow c+xw$}
\ENDFOR
\STATE Write $Y_{ij}\leftarrow c$.
\ENDFOR
\RETURN $Y$
\end{algorithmic}
\end{minipage}
\end{algorithm}

%% file: sections/appendix_interventions.tex
\section{Additional intervention results}
\label{sec:supporting-results}

\autoref{fig:discussion-summary} summarizes every intervention against \targetbase{} and every distilled student against the NVIDIA-teacher student.
\input{figures/fig_discussion_summary}

\subsection{Complete output comparison}
\label{sec:detail-correction-results}
\autoref{tab:head-results-full} extends \autoref{tab:head-results} with probability mean squared error (MSE), top-1 agreement, and top-5 set agreement.
\input{tables/main_results_full}

\newpage

\subsection{Loss weighting}
\label{sec:detail-loss-balancing}
\label{sec:loss-balancing}
\begin{wraptable}{r}{0.36\linewidth}
\vspace{-\intextsep}
\centering
\caption{Loss weights after normalization.}
\label{tab:loss-weights}
\scriptsize
\setlength{\tabcolsep}{3pt}
\renewcommand{\arraystretch}{1.05}
\begin{tabular}{@{}lrrrr@{}}
\hline
Profile & MSE & KL & Token & Top-5 \\
\hline
\profileEqual{} & 0.25 & 0.25 & 0.25 & 0.25 \\
\profileMSE{} & 0.70 & 0.10 & 0.10 & 0.10 \\
\profileKL{} & 0.10 & 0.70 & 0.10 & 0.10 \\
\profileToken{} & 0.10 & 0.10 & 0.70 & 0.10 \\
\profileTopFive{} & 0.10 & 0.10 & 0.10 & 0.70 \\
\profileNumerical{} & 0.40 & 0.40 & 0.10 & 0.10 \\
\profileDecision{} & 0.10 & 0.10 & 0.40 & 0.40 \\
\hline
\end{tabular}
\vspace{-\intextsep}
\end{wraptable}
A single weighting may emphasize the wrong aspect of agreement, so we train seven profiles that change only the loss coefficients.
\autoref{tab:loss-weights} lists the seven weighting profiles, \autoref{fig:loss-weighting-tradeoffs} places their numerical and choice agreement together, and \autoref{tab:head-weighting-results} gives all seven profiles on the measures of the main table. No profile lowers raw-logit MSE or forward KL below \targetbase{}. \profileKL{} has the lowest MSE among them, \profileNumerical{} lowers MSE relative to \profileEqual{} while raising KL, and \profileDecision{} gains one rollout while losing ordered top-5 agreement. A profile's name therefore describes what training emphasizes rather than what improves. On OLMoE, every profile raises both MSE ($+0.25$ to $+0.65\%$) and KL ($+2.8$ to $+5.3\%$), with all paired intervals above zero, and matches 486 to 1,211 fewer ordered top-5 lists. \profileMSE{} gains a rollout (45 of 128, against 44) while its numerical error grows. Every OLMoE profile ends above \targetbase{} on MSE and KL, leaves expert selection unchanged, and moves benchmark counts by at most one answer.

\input{figures/fig_loss_weighting}
\input{tables/head_weighting_results}

\input{sections/head_capacity}

%% file: figures/fig_discussion_summary.tex
\begin{figure}[t]
    \centering
    \includegraphics[width=\linewidth]{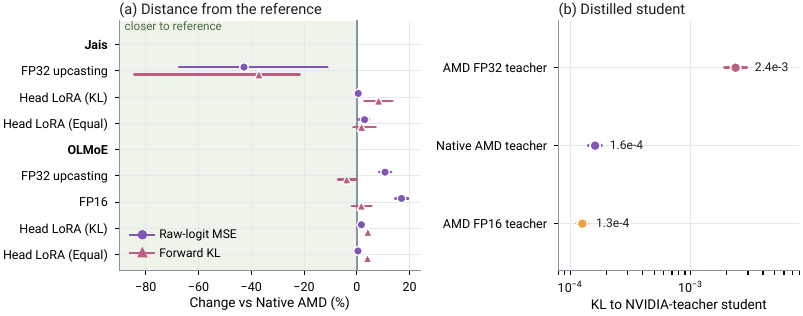}
    \caption{(a) Change in raw-logit MSE and forward KL relative to \targetbase{}. Values left of zero are closer to the NVIDIA reference, and Jais FP16 is omitted because it produces nonfinite outputs. Only FP32 upcasting on Jais moves toward the reference on both measures, and every other intervention moves away on at least one. (b) Forward KL of each distilled student from the NVIDIA-teacher student.}
    \label{fig:discussion-summary}
\end{figure}

%% file: tables/main_results_full.tex
\begin{table}[t]
\centering
\caption{Complete output comparison for \autoref{tab:head-results}. Choices use 2,498 positions and 200 rollouts for Jais and 131,072 positions and 128 rollouts of 1,024 tokens for OLMoE. The greedy next token on the reference input equals Top-1 and is not repeated. Jais accuracy parses generated answers on 13,890 MMLU and 14,303 ArabicMMLU questions, including the adapter's training questions (\autoref{tab:accuracy-no-overlap} excludes them), and OLMoE accuracy scores answer-letter logits on the full MMLU and ARC-Challenge test sets. -- marks non-BF16 outputs.}
\label{tab:head-results-full}
\footnotesize
\setlength{\tabcolsep}{2.2pt}\renewcommand{\arraystretch}{1.1}
\begin{tabular}{@{}lrrrrrrrrrr@{}}
\hline
& \multicolumn{4}{c}{Numerical values} & \multicolumn{4}{c}{Choices (\%) $\uparrow$} & \multicolumn{2}{c}{Accuracy (\%) $\uparrow$} \\
\cline{2-5}\cline{6-9}\cline{10-11}
Method & \shortstack{Bit-eq.\\(\%) $\uparrow$} & \shortstack{Logit\\MSE $\downarrow$} & \shortstack{Prob.\ MSE\\($10^{-7}$) $\downarrow$} & \shortstack{KL\\($10^{-3}$) $\downarrow$} & Top-1 & \shortstack{Ord.\\top-5} & \shortstack{Top-5\\set} & Rollout & \shortstack{MMLU} & \shortstack{AMMLU/\\ARC} \\
\hline
\multicolumn{11}{@{}l}{\textit{Jais}} \\
\targetbase{} & 9.26 & 0.2873 & 0.279 & 7.08 & 98.76 & 66.53 & 83.91 & 93.50 & 60.86 & 71.36 \\
FP16 & \multicolumn{10}{c}{nonfinite outputs} \\
FP32 upcasting & -- & 0.1644 & 0.186 & 4.45 & 98.68 & 71.58 & 86.55 & 93.00 & 61.00 & 71.40 \\
Head LoRA (KL) & 9.22 & 0.2888 & 0.298 & 7.67 & 98.52 & 65.89 & 83.23 & 92.50 & 60.88 & 71.35 \\
Head LoRA (\profileEqual{}) & 8.65 & 0.2957 & 0.278 & 7.21 & 98.64 & 66.37 & 83.23 & 93.50 & 60.87 & 71.34 \\
\hline
\multicolumn{11}{@{}l}{\textit{OLMoE}} \\
\targetbase{} & 14.17 & 0.0215 & 0.00833 & 0.194 & 99.85 & 69.57 & 85.10 & 34.38 & 47.37 & 54.10 \\
FP16 & -- & 0.0251 & 0.00796 & 0.198 & 99.84 & 66.66 & 83.42 & 35.16 & 47.46 & 54.10 \\
FP32 upcasting & -- & 0.0238 & 0.00759 & 0.187 & 99.84 & 67.75 & 84.04 & 33.59 & 47.44 & 54.69 \\
Head LoRA (KL) & 13.03 & 0.0219 & 0.00863 & 0.202 & 99.85 & 69.00 & 84.87 & 33.59 & 47.37 & 54.10 \\
Head LoRA (\profileEqual{}) & 13.92 & 0.0216 & 0.00866 & 0.202 & 99.85 & 69.04 & 84.83 & 33.59 & 47.37 & 54.10 \\
\hline
\end{tabular}
\end{table}

%% file: figures/fig_loss_weighting.tex
\begin{figure}[t]
    \centering
    \includegraphics[width=\linewidth]{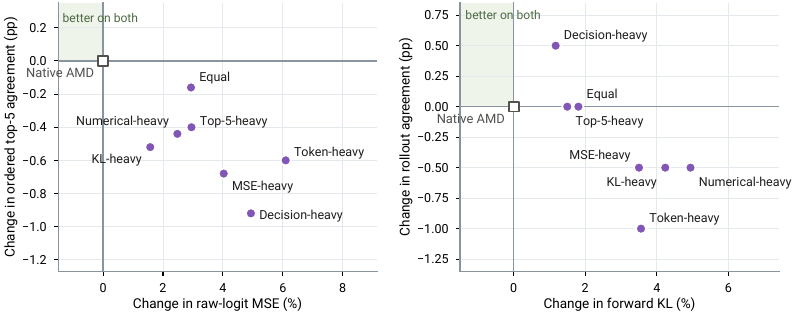}
    \caption{Effect of loss weighting on Jais at the fixed training endpoint, relative to \targetbase{} (open square). The left panel plots the change in raw-logit mean squared error (MSE) against change in ordered top-5 agreement over 2,498 positions. The right panel plots the change in forward Kullback--Leibler (KL) divergence against change in rollout agreement over 200 responses, where one response is 0.5 percentage points. The shaded region would improve both measures. No weighting profile reaches it in the left panel, and only \profileDecision{} gains a response on the right while still increasing KL.}
    \label{fig:loss-weighting-tradeoffs}
\end{figure}

%% file: tables/head_weighting_results.tex
\begin{table}[t]
\centering
\caption{Loss-weighting comparison at the fixed training endpoint. Positions, rollouts, and accuracy sets are as in \autoref{tab:head-results-full}, and the last column is ArabicMMLU for Jais and ARC-Challenge for OLMoE. Jais accuracy includes the adapter's training questions (\autoref{tab:accuracy-no-overlap} excludes them).}
\label{tab:head-weighting-results}
\begin{draftresulttable}
\begin{tabular}{@{}lrrrrrrr@{}}
\hline
Method & \shortstack{Bit-eq.\\(\%) $\uparrow$} & \shortstack{Logit\\MSE $\downarrow$} & \shortstack{KL\\($10^{-3}$) $\downarrow$} & \shortstack{Ord.\ top-5\\(\%) $\uparrow$} & \shortstack{Rollout\\(\%) $\uparrow$} & \shortstack{MMLU\\(\%) $\uparrow$} & \shortstack{AMMLU/\\ARC (\%) $\uparrow$} \\
\hline
\multicolumn{8}{@{}l}{\textit{Jais}} \\
\targetbase{} & 9.26 & 0.2873 & 7.08 & 66.53 & 93.50 & 60.86 & 71.36 \\
\profileEqual{} & 8.65 & 0.2957 & 7.21 & 66.37 & 93.50 & 60.87 & 71.34 \\
\profileMSE{} & 8.72 & 0.2988 & 7.33 & 65.85 & 93.00 & 60.89 & 71.36 \\
\profileKL{} & 8.95 & 0.2918 & 7.38 & 66.01 & 93.00 & 60.89 & 71.34 \\
\profileToken{} & 8.07 & 0.3048 & 7.33 & 65.93 & 92.50 & 60.89 & 71.36 \\
\profileTopFive{} & 8.65 & 0.2957 & 7.19 & 66.13 & 93.50 & 60.96 & 71.32 \\
\profileNumerical{} & 8.75 & 0.2944 & 7.43 & 66.09 & 93.00 & 60.86 & 71.34 \\
\profileDecision{} & 8.37 & 0.3014 & 7.16 & 65.61 & 94.00 & 60.91 & 71.35 \\
\hline
\multicolumn{8}{@{}l}{\textit{OLMoE}} \\
\targetbase{} & 14.17 & 0.02149 & 0.194 & 69.57 & 34.38 & 47.37 & 54.10 \\
\profileEqual{} & 13.92 & 0.02158 & 0.202 & 69.04 & 33.59 & 47.37 & 54.10 \\
\profileMSE{} & 14.02 & 0.02154 & 0.200 & 69.20 & 35.16 & 47.37 & 54.10 \\
\profileKL{} & 13.77 & 0.02163 & 0.202 & 69.08 & 32.81 & 47.37 & 54.10 \\
\profileToken{} & 13.96 & 0.02157 & 0.203 & 68.96 & 32.81 & 47.37 & 54.10 \\
\profileTopFive{} & 13.88 & 0.02160 & 0.205 & 68.65 & 32.81 & 47.37 & 54.10 \\
\profileNumerical{} & 13.93 & 0.02158 & 0.201 & 69.14 & 32.81 & 47.37 & 54.10 \\
\profileDecision{} & 13.92 & 0.02159 & 0.204 & 68.86 & 34.38 & 47.37 & 54.10 \\
\hline
\end{tabular}
\end{draftresulttable}
\end{table}

%% file: sections/head_capacity.tex
\subsection{Capacity of an output-head correction}
\label{sec:head-capacity}\label{sec:lora-not-enough}
\begin{wraptable}{r}{0.40\linewidth}
\vspace{-\intextsep}
\centering
\caption{Share of squared Jais logit mismatch a linear head map can remove, same inputs.}
\label{tab:head-capacity}
\footnotesize
\setlength{\tabcolsep}{3pt}
\renewcommand{\arraystretch}{1.05}
\begin{tabular}{@{}lr@{}}
\hline
Map & Removed \\
\hline
Rank 32 & 11.87\% \\
Any rank (all directions) & 26.09\% \\
\hline
Outside any linear map & 73.91\% \\
\hline
\end{tabular}
\vspace{-\intextsep}
\end{wraptable}
Before training any adapter, we ask how much of the Jais mismatch a shared linear correction at the output head could represent at all. Let $H\in\mathbb{R}^{N\times d}$ hold the AMD hidden values entering the head and $E\in\mathbb{R}^{N\times V}$ the NVIDIA-minus-AMD logit residuals at the same $N=19{,}316$ training positions, with $d=3{,}328$ and $V=150{,}272$. A rank-$r$ head correction has the form $HW$ with $\operatorname{rank}(W)\le r$. \autoref{tab:head-capacity} gives the fraction of $\lVert E\rVert_F^2$ that the best such map removes. Even an ideal rank-32 map removes 11.9\%, and using every hidden direction removes only 26.1\%. These are in-sample fits, and a regression on pure noise with all 3,328 directions would already fit about 17\% ($d/N$), so the out-of-fold result below is the relevant one.

With $G=H^\top H=U\operatorname{diag}(\lambda)U^\top$ and $R=H^\top E$, we form $K=\operatorname{diag}(\lambda)^{-1/2}U^\top R$ over the retained directions and $T=KK^\top$. The top $r$ eigenvalues of $T$ divided by $\lVert E\rVert_F^2$ give the rank-$r$ fraction, and its trace gives the all-direction fraction. All sums use FP64, positions have equal weight, and no intercept is used. The cutoff $\tau=\max(N,d)\,\epsilon_{64}\,\lambda_{\max}(G)$ keeps 3,327 of 3,328 directions, and keeping all of them changes both fractions by less than 0.01 percentage points.

These are upper bounds on the same inputs rather than results of a trained adapter. On new prompts the representable part does not transfer either, as regularized rank-32 maps fitted on disjoint folds increase the pooled out-of-fold squared error, by 0.81\% in the best setting. The analysis does not establish the rank of $E$ itself, since a residual can have concentrated structure that the hidden value still cannot predict.

%% file: sections/appendix_extensions.tex
\section{Additional experiments}
\label{sec:additional-experiments}

\subsection{Two further models}
\label{sec:extension-models}
Output-head LoRA fails in both Jais and OLMoE, so these experiments focus on the precision interventions. We repeat the OLMoE protocol on a second dense model, Llama-3.1-8B-Instruct~\citep{grattafiori2024llama3herdmodels}, and a second MoE model, Qwen1.5-MoE-A2.7B-Chat~\citep{qwen2024qwen15moe}, which routes each token to four of 60 experts. Both use the same 128 development responses with the 1,024-token cap, the NVIDIA A100 BF16 reference, and the AMD MI210 target with byte-identical weights. Logit measures use 32 of these responses at every position, and rollouts use all 128. \autoref{tab:extension-models} shows that both models reproduce the pattern of the main study. The MoE model again has the larger mismatch, as only one of its 128 responses matches the reference and the four routed experts agree at every layer for only 16.9\% of positions. FP32 upcasting lowers the logit error of Llama but makes fewer of its responses identical, and in Qwen1.5-MoE both finer formats move every logit and router measure away from the reference, as they do in OLMoE. Accuracy stays within a few dozen answers across formats and vendors in both models, so the change in behavior again remains invisible to accuracy.

\begin{table}[ht]
\centering
\caption{Two further models on AMD against the NVIDIA BF16 reference. Router is the share of positions and layers where the four routed experts match. Accuracy is on the full MMLU and ARC-Challenge test sets. -- marks non-BF16 outputs, a dense model, or the reference itself.}
\label{tab:extension-models}
\begin{draftresulttable}
\begin{tabular}{@{}lrrrrrrrr@{}}
\hline
& Bit-equal & Logit & KL & Top-5 & Router & Rollout & MMLU & ARC \\
Method & (\%)$\uparrow$ & MSE$\downarrow$ & ($10^{-3}$)$\downarrow$ & (\%)$\uparrow$ & (\%)$\uparrow$ & (\%)$\uparrow$ & (\%) & (\%) \\
\hline
\multicolumn{9}{@{}l}{\textit{Llama-3.1-8B-Instruct (dense)}} \\
NVIDIA reference & -- & -- & -- & -- & -- & -- & 66.93 & 83.45 \\
\targetbase{} & \textbf{9.52} & 0.00533 & 0.218 & 80.27 & -- & \textbf{16.41} & 67.04 & 83.62 \\
FP32 upcasting & -- & \textbf{0.00449} & \textbf{0.130} & \textbf{81.00} & -- & 8.59 & 67.06 & 83.45 \\
FP16 & -- & 0.00586 & 0.151 & 79.33 & -- & 10.94 & 67.06 & 83.53 \\
\hline
\multicolumn{9}{@{}l}{\textit{Qwen1.5-MoE-A2.7B-Chat (MoE)}} \\
NVIDIA reference & -- & -- & -- & -- & -- & -- & 60.48 & 76.71 \\
\targetbase{} & \textbf{5.64} & \textbf{0.01519} & \textbf{1.66} & \textbf{66.75} & \textbf{87.93} & 0.78 & 60.74 & 76.54 \\
FP32 upcasting & -- & 0.01772 & 2.01 & 63.89 & 85.65 & \textbf{2.34} & 60.84 & 76.62 \\
FP16 & -- & 0.01991 & 2.30 & 62.22 & 84.77 & \textbf{2.34} & 60.92 & 76.79 \\
\hline
\end{tabular}
\end{draftresulttable}
\end{table}

\subsection{Matched precision on both vendors}
\label{sec:matched-format}
The precision interventions of \autoref{sec:precision-results} change the format on AMD alone, so the target no longer rounds the way the BF16 reference does. We therefore also run the NVIDIA reference forward in FP32 upcasting and in FP16 over the same OLMoE reference responses and compare each AMD format with the NVIDIA run of the same format. \autoref{tab:matched-format} shows the result on 32 development responses, with rollouts over all 128. With FP32 on both vendors, the outputs agree almost exactly, and 98.4\% of the responses become identical. FP16 on both vendors lies between this and the BF16 pair. The mismatch in the main study is therefore largely a property of BF16 execution, and a shared higher-precision format removes most of it at the cost measured in \autoref{tab:precision-cost}.

\begin{table}[ht]
\centering
\caption{OLMoE with the same or a different format on each vendor. Router is the share of positions and layers where the eight selected experts match.}
\label{tab:matched-format}
\begin{draftresulttable}
\begin{tabular}{@{}llrrrrr@{}}
\hline
& & Logit MSE & KL & Top-5 & Router & Rollout \\
AMD & NVIDIA & ($10^{-3}$)$\downarrow$ & ($10^{-4}$)$\downarrow$ & (\%)$\uparrow$ & (\%)$\uparrow$ & (\%)$\uparrow$ \\
\hline
BF16 & BF16 & 18.9 & 2.17 & 70.69 & 90.40 & 34.38 \\
FP16 & BF16 & 22.6 & 2.15 & 68.14 & 87.87 & 35.16 \\
FP32 upcasting & BF16 & 21.4 & 2.06 & 69.03 & 88.38 & 33.59 \\
\hline
FP16 & FP16 & 14.9 & 0.882 & 78.40 & 93.25 & 60.16 \\
FP32 upcasting & FP32 upcasting & \textbf{0.086} & \textbf{0.0019} & \textbf{99.85} & \textbf{99.97} & \textbf{98.44} \\
\hline
\end{tabular}
\end{draftresulttable}
\end{table}

\subsection{Selective FP32 in Jais}
\label{sec:selective-fp32}
FP32 upcasting of the whole Jais forward pass gives the largest numerical gain in the main study but triples the runtime. We therefore compute only chosen modules in FP32, casting the BF16 weights losslessly, and evaluate on the same development responses and reference. \autoref{tab:selective-fp32} and \autoref{fig:selective-fp32} show that the gain depends strongly on where the higher precision acts. When each module rounds its output back to BF16, only LayerNorm and attention lower both MSE and KL, the output head changes almost nothing, and the MLPs raise KL while being the most expensive module type to upcast. Combining LayerNorm and attention keeps more than half of the full KL reduction, and adding the output head or the last eight blocks adds little, because the residual stream between the modules is still rounded to BF16.

We therefore keep everything in FP32 except the MLPs, including the residual stream, and report this setting as selective FP32 in \autoref{tab:head-results}. It recovers more than 90\% of the full reduction in raw-logit MSE and forward KL, matches almost as many ordered top-5 lists, and keeps the most responses identical to the reference, at 1.30 times the forward time of \targetbase{} against 3.02 times for full upcasting. Its MMLU and ArabicMMLU accuracy stays within 16 answers of \targetbase{}. The residual stream therefore carries most of the rounding that full upcasting removes, while the MLPs can stay in BF16. For settings whose output head runs in FP32, \autoref{tab:selective-fp32} counts bit-equal logits after rounding them to BF16. The FP32 output head alone then reproduces the \targetbase{} logits exactly, so the head's own arithmetic adds nothing to the mismatch.

\input{figures/fig_selective_fp32}

\begin{table}[!ht]
\providecommand{\secondbest}[1]{\raisebox{0pt}[\height][0pt]{\underline{#1}}}
\centering
\caption{Jais with FP32 computation in selected modules}
\label{tab:selective-fp32}
\begin{draftresulttable}
\begin{tabular}{@{}lrrrrrr@{}}
\hline
FP32 modules & \shortstack{Bit-equal\\(\%)$\uparrow$} & \shortstack{Logit\\MSE$\downarrow$} & \shortstack{KL\\($10^{-3}$)$\downarrow$} & \shortstack{Top-5\\(\%)$\uparrow$} & \shortstack{Rollout\\(\%)$\uparrow$} & \shortstack{Time\\($\times$)$\downarrow$} \\
\hline
\targetbase{} & 9.26 & 0.2873 & 7.08 & 66.53 & \secondbest{93.50} & \textbf{1.00} \\
\hline
\multicolumn{7}{@{}l}{\textit{Single module type}} \\
Output head & 9.26 & 0.2865 & 6.87 & 66.21 & \secondbest{93.50} & \secondbest{1.01} \\
LayerNorm & 9.20 & 0.2522 & 5.81 & 66.89 & \secondbest{93.50} & 1.02 \\
Attention & 10.00 & 0.2077 & 6.31 & 68.73 & 91.50 & 1.28 \\
MLP & 9.31 & 0.2449 & 7.52 & 66.57 & 91.00 & 2.10 \\
\hline
\multicolumn{7}{@{}l}{\textit{Depth slice}} \\
First 8 blocks & 9.40 & 0.2441 & 8.37 & 67.25 & 91.00 & 1.35 \\
Last 8 blocks & 9.59 & 0.2726 & 6.62 & 67.45 & \secondbest{93.50} & 1.36 \\
\hline
\multicolumn{7}{@{}l}{\textit{Combination}} \\
LayerNorm + attention & 10.01 & 0.2142 & 5.54 & 67.73 & 90.50 & 1.30 \\
LayerNorm + attention + head & 10.01 & 0.2135 & 5.46 & 67.77 & 92.00 & 1.31 \\
LayerNorm + attention + last 8 & 10.12 & 0.2096 & 5.80 & 68.94 & 92.00 & 1.93 \\
All except MLP (selective FP32) & \secondbest{11.07} & \secondbest{0.1717} & \secondbest{4.62} & \secondbest{71.38} & \textbf{94.00} & 1.30 \\
All (FP32 upcasting) & \textbf{11.24} & \textbf{0.1644} & \textbf{4.45} & \textbf{71.58} & 93.00 & 3.02 \\
\hline
\end{tabular}
\end{draftresulttable}
\end{table}

%% file: figures/fig_selective_fp32.tex
\begin{figure}[!ht]
    \centering
    \includegraphics[width=\linewidth]{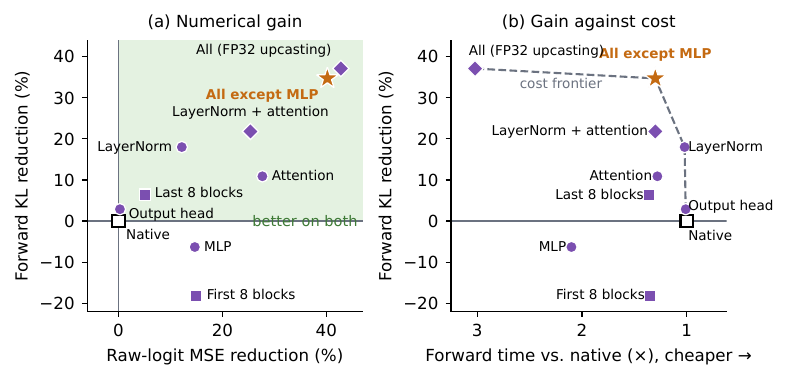}
    \caption{Selective FP32 settings on Jais, relative to \targetbase{} (open square). (a) Reduction in raw-logit MSE and forward KL, where the shaded region improves both. (b) KL reduction against forward time, with cheaper settings to the right. The dashed line joins the settings that no cheaper setting beats on KL. Keeping everything except the MLPs in FP32 (star) lies on this frontier near full upcasting at 1.30 times the forward time.}
    \label{fig:selective-fp32}
\end{figure}

%% file: sections/limitations.tex
\section{Limitations}
\label{sec:limitations}
We compare one pair of accelerators, the NVIDIA A100 and the AMD MI210, under one software stack with PyTorch 2.7.1, CUDA 12.8, and ROCm 6.3.3. Other GPU generations, vendor libraries, attention kernels, and batch sizes change the kernels that run and can change the size of the mismatch. The exact magnitudes we report therefore apply to this configuration, while the mechanism and its propagation through the model do not depend on it.

We study four open models of up to 14 billion parameters on multiple-choice benchmarks. Longer free-form generation, larger models, and other task types may accumulate more mismatch than we observe. We also treat the NVIDIA execution as an operational reference and do not compare either vendor with a higher-precision ground truth.